\pdfoutput=1

\documentclass[11pt]{article}

\usepackage[preprint]{acl}

\usepackage{times}
\usepackage{latexsym}

\usepackage[T1]{fontenc}

\usepackage[utf8]{inputenc}

\usepackage{microtype}

\usepackage{inconsolata}

\usepackage{natbib}

\usepackage{graphicx}
\usepackage{algorithm}
\usepackage{algorithmic}
\usepackage{amsmath}
\usepackage{amsfonts}
\usepackage{multirow}
\usepackage{xcolor}
\definecolor{darkgreen}{rgb}{0, 0.7, 0} 
\definecolor{darkred}{rgb}{0.9, 0, 0}   
\usepackage{enumerate}
\usepackage{float}  
\usepackage{subfigure}  
\usepackage{url}
\usepackage{lipsum} 
\usepackage{array}
\usepackage{colortbl}
\usepackage{pifont}
\usepackage{CJKutf8}

\usepackage{xurl}

\title{NatLan: Native Language Prompting Facilitates Knowledge Elicitation Through Language Trigger Provision and Domain Trigger Retention}

\author{Baixuan Li$^{1}$ \qquad Yunlong Fan$^{1}$ \qquad Tianyi Ma$^{2}$ \qquad
  Zhiqiang Gao$^{* 1}$\\
  $^1$School of Computer Science and Engineering, Southeast University, Nanjing 211189, China\\
  $^2$Department of Computer Science and Engineering, Michigan State University\\
  \texttt{\{baixuan, fanyunlong, zqgao\}@seu.edu.cn }\\
  \texttt{matiany3@msu.edu}\\
  }

\begin{document}
\maketitle


\begin{abstract}

Multilingual large language models (MLLMs) do not perform as well when answering questions in non-dominant languages as they do in their dominant languages. Although existing translate-then-answer methods alleviate this issue, the mechanisms behind their effectiveness remain unclear. In this study, we analogize the dominant language of MLLMs to the native language of humans and use two human cognitive features: \textbf{the Language Trigger} (\textbf{LT}) and \textbf{the Domain Trigger} (\textbf{DT}), to interpret the mechanisms behind translate-then-answer methods. This reveals that while sufficient LTs are provided by these methods, there remains a deficiency in DT retention. To mitigate this issue, we propose \textbf{Nat}ive \textbf{Lan}guage Prompting (\textbf{NatLan}), employing a Multi-MLLM collaboration strategy and introducing an additional role-enhanced domain-specific MLLM with stronger multilingual understanding capabilities as the translator. Across five language QA benchmarks, NatLan achieves up to a \textbf{31.28\%} improvement in accuracy and, compared to existing state-of-the-art methods, provides comparable or greater retention of DTs in up to \textbf{87\%} of cases.
Our code is available at \url{https://github.com/AnonyNLP/NatLan}.
\end{abstract}

\section{Introduction}

\begin{figure}[ht] 
\centering
\includegraphics[width=1\linewidth]{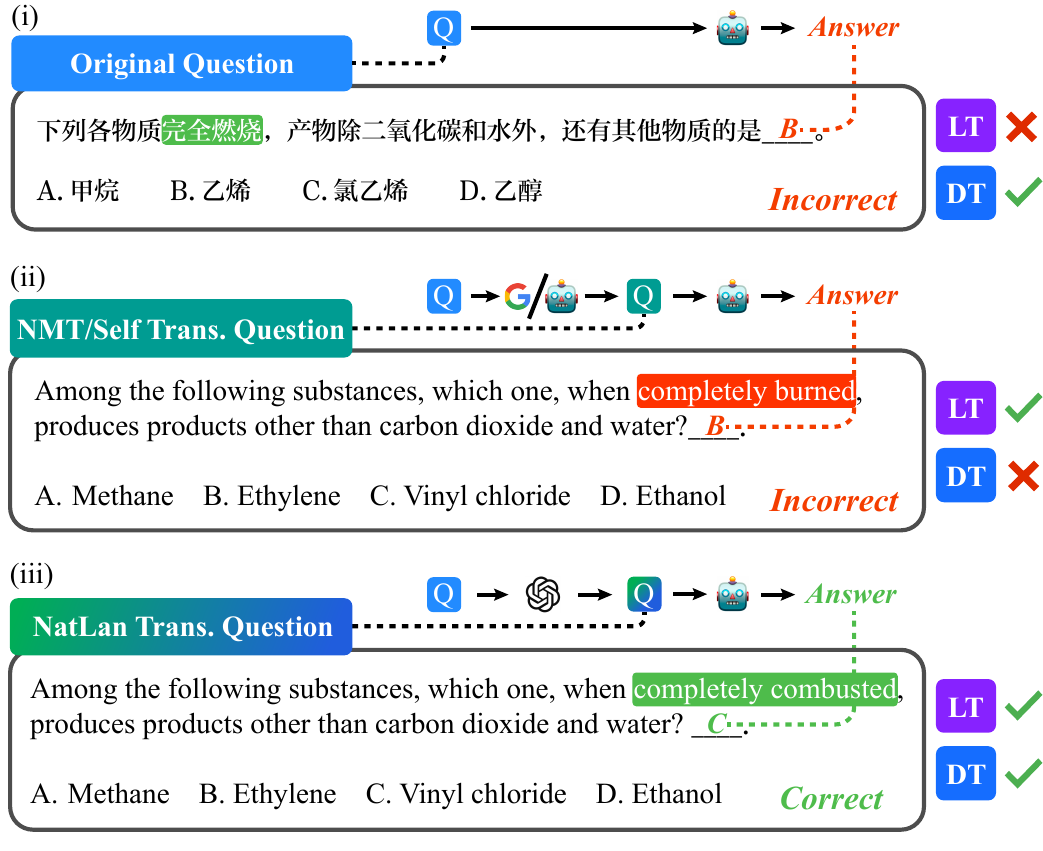}
\caption{
The presence of Language Triggers (LTs) and Domain Triggers (DTs) in questions processed by different methods when addressing non-native language QA. The same icon represents the same question/model.
}
\label{intro}
\end{figure}

Recent research has observed that multilingual large language models (MLLMs) \cite{brown2020language, achiam2023gpt} fail to answer certain questions in non-dominant languages that they can correctly answer when posed in their dominant languages (i.e., the language with the highest proportion during training\footnote{Such as English for Llama \cite{touvron2023llama}, which accounts for over 70\% of the tokens in the pretraining corpus.}) \cite{zhang2023don, huang2023not, etxaniz2024multilingual}. 
Currently, two main \textit{translate-then-answer} \cite{schulhoff2024prompt} methods are employed to resolve this issue. Both involve translating the question into the MLLM's dominant language before it answers, thereby facilitating the use of the MLLM's richer dominant language knowledge. One method allows the MLLM itself to handle the translation \cite{zhang2023don, huang2023not, etxaniz2024multilingual}, while the other uses an external Neural Machine Translation (NMT) system \cite{shi2022language}. We refer to the former as \textit{Self-Translation} and the latter as \textit{NMT-Translation}.
However, what are the mechanisms behind the effectiveness of translate-then-answer methods, and how do they affect the elicitation of dominant language knowledge in MLLMs?

To investigate these issues, we analogize the dominant language of MLLMs to the native language of humans and interpret the mechanisms of the translate-then-answer methods through two key features observed in human cognitive processes, which we respectively term as \textbf{Language Trigger (LT)} and \textbf{Domain Trigger (DT)}. The former reduces the cognitive load of understanding the non-native questions by translating them into the native language \cite{wu2022cognitive, zeng2022first, gao2023shared, del2022decision}, while the latter narrows the scope of knowledge elicitation by employing domain-specific terms \cite{baker2015role, acheampong2016answer, tong2020improving, liu2022saliency}. Together, these cognitive triggers facilitate the elicitation of knowledge when humans answer non-native language questions.

Accordingly, we analyze the occurrences of these cognitive triggers in existing methods. 
As depicted in Figure \ref{intro} (i), when the MLLM directly answers a question in the non-native language (Chinese), it lacks appropriate LTs. Even if correct DTs are present in the human-constructed original question, the method fails to accurately elicit the knowledge within the MLLM. 
As depicted in Figure \ref{intro} (ii), although the existing translate-then-answer methods are capable of translating and providing sufficient LTs (questions in the native language, English) for the MLLM, their inherent deficiencies lead to inadequate translation of fine-grained specialized terms (DTs) \cite{tu2017neural, zhu2023investigating, ai2023tecs}, translating the specialized chemical term ``completely combusted'' as the more generic ``completely burned''. 
Consequently, the absence of DTs leads to the failure of the relevant knowledge elicitation in MLLMs.

To mitigate this issue observed in existing translate-then-answer methods, we propose \textbf{Nat}ive \textbf{Lan}guage Prompting (\textbf{NatLan}), which employs a role-enhanced domain-specific Multi-MLLM collaboration strategy \cite{talebirad2023multi, dong2024self}, comprising the Translator LLM and the Speaker LLM. The former translates questions from a non-native language to the native language of the Speaker LLM, while the latter answers questions based on the translated questions. As depicted in Figure \ref{intro} (iii), as a novel translate-then-answer method, NatLan provides sufficient LTs while maximally retaining DTs during translation, further facilitating the elicitation of relevant knowledge during non-native language QA.

Our contributions are primarily as follows:
\textbf{(i) Cognition-inspired Interpretation:} We employ two cognitive triggers, LTs and DTs, to interpret the limitations of existing translate-then-answer methods in non-native language QA as the insufficiency of DTs.
\textbf{(ii) Effective Remediation:} We propose NatLan to mitigate this issue, achieving up to a \textbf{31.28\%} improvement in accuracy across five non-native language QA benchmarks and surpassing all top-notch methods.
\textbf{(iii) Multi-level Confirmation:} We confirmed that NatLan provides comparable or greater retention of DTs in up to \textbf{87\%} of cases at the input level compared to the state-of-the-art NMT-Translation method, and further demonstrate at the activation level that a higher number of DTs results in more salient knowledge activation, which helps correct the MLLM's answers. This validates the rationality of using cognitive triggers to interpret the translate-then-answer process. 

\section{Related Work}

\paragraph{Language Triggers and Domain Triggers in Cognitive Processes.}
For human multilinguals, there are two key features that assist in leveraging their native language knowledge to correctly answer questions in a less proficient non-native language. The first one aligns with the human tendency to prioritize thinking in the native \textbf{language} by converting non-native language questions into the corresponding native language \cite{wu2022cognitive, zeng2022first, gao2023shared}, thereby reducing the cognitive load associated with understanding the question \cite{del2022decision}. The second one aligns with the fact that different \textbf{domains} of knowledge have their own specialized terminology. Appropriate use of domain-specific terms enables humans to more easily associate knowledge within that field \cite{baker2015role, acheampong2016answer, tong2020improving, liu2022saliency}, thereby narrowing the scope of knowledge elicitation to a specific domain with greater precision.

In this study, we refer to these cognitive features as the \textbf{Language Trigger} (\textbf{LT}) and the \textbf{Domain Trigger} (\textbf{DT}), respectively. Subsequently, we use them to interpret translate-then-answer methods and propose NatLan to address the limitations of existing methods with respect to these two triggers.

\begin{figure*}[ht] 
\centering
\includegraphics[width=1\linewidth]{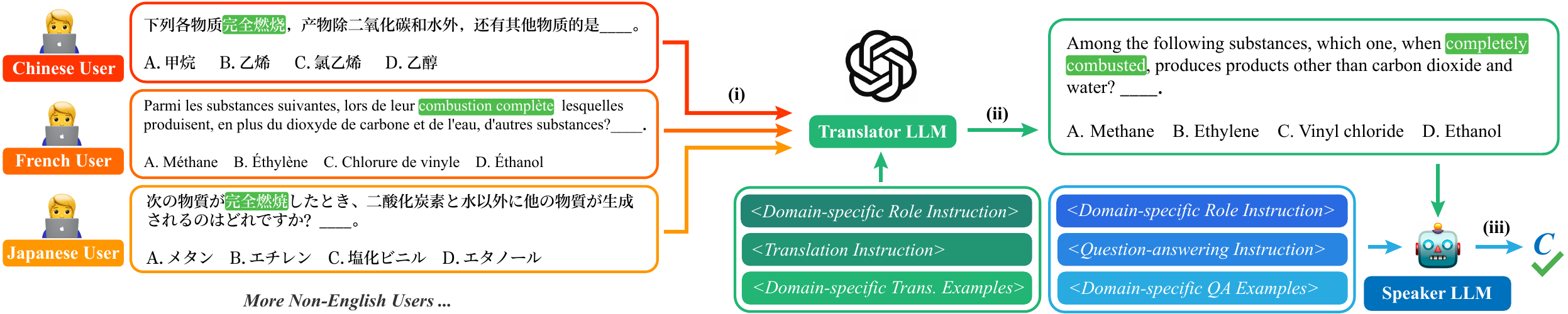}
\caption{Non-native language question-answering workflow of NatLan. (i) Non-English users issue queries. (ii) The Translator LLM translates the non-native language questions into the native language (English) of the Speaker LLM. (iii) The Speaker LLM answers the native language question. More details are available in Appendix \ref{appendix:details}.}
\label{fig:method}
\end{figure*}

\paragraph{Translate-then-answer Prompting.}
Translate-then-answer Prompting \cite{schulhoff2024prompt} aims to leverage the strength of MLLMs in their dominant/native language (English). One category of these methods falls under Self-Translation \cite{zhang2023don, huang2023not, etxaniz2024multilingual}, which requires MLLMs themselves to perform the non-native to native language translation, before answering questions. However, if the model has poor capabilities in the non-native language, it may not capture the DTs in the original questions accurately. \citet{shi2022language} used external Neural Machine Translation (NMT) systems to translate the questions. However, unlike MLLMs \cite{vilar2023prompting, guo2024teaching, kang2024translate}, NMT systems are translation-only and lack domain-specific knowledge, resulting in overly generic and literal translations \cite{tu2017neural, zhu2023investigating, ai2023tecs}.  Overall, while existing methods can provide sufficient LTs through translation, they inevitably lose some of the DTs in the original questions during the translation process, leading to suboptimal or even detrimental effects on the subsequent answers of MLLMs.

Our proposed NatLan incorporates an additional role-enhanced domain-specific MLLM with stronger multilingual capabilities as the translator. This approach mitigates the limitations of Self-Translation, which is constrained by its own capabilities, and NMT-Translation, which lacks sufficient domain knowledge. As a result, it ensures the provision of LTs while maximizing the retention of DTs during the translate-then-answer process.

\section{Role-Enhanced Multi-MLLM Collaboration}
\label{sec:collaboration}

Since the capabilities of a single MLLM are limited, and different MLLMs exhibit varying strengths, in order to allow each MLLM to fully leverage its unique advantages, previous work has proposed using multiple MLLMs to fulfill distinct roles within a collaborative framework \cite{talebirad2023multi, dong2024self}. In this study, the translate-then-answer process is inherently divided into two sub-processes: (i) translating and (ii) answering. Accordingly, we designed a Role-Enhanced Multi-MLLM Collaboration framework and defined two distinct roles to handle these sub-processes separately. Their respective targets and required characteristics are outlined as follows:

\begin{enumerate}[(i)]
\item 
\textbf{Translator} requires MLLMs to have strong multilingual comprehension and semantic preservation abilities. It needs to translate any received non-native language questions into the native language of the Speaker LLM.

\item 
\textbf{Speaker} requires MLLMs that excel in their native language (dominant language) and are capable of understanding the given non-native language, though not necessarily to an exceptional degree. It needs to rely on its own knowledge to provide answers to the questions translated by the Translator LLM.

\end{enumerate}


\section{Native Language Prompting}
\label{sec:natlan}
Utilizing our constructed Role-Enhanced Multi-MLLM Collaboration framework, we further proposed \textbf{Nat}ive \textbf{Lan}guage Prompting (\textbf{NatLan}) to provide rich Language Triggers (LTs) while maximally retaining Domain Triggers (DTs) during the translate-then-answer process.

As depicted in Figure \ref{fig:method}, in addition to constructing the domain-specific role instruction, we also injected domain-specific context through few-shot (5-shot) examples to achieve two objectives: \textbf{(i) Individual Enhancement:} facilitating the recall of domain-specific knowledge by each MLLM in this framework through domain-specific role prompting. \textbf{(ii) Joint Enhancement:} maximizing the retention of DTs by the Translator LLM during the translation process, allowing the rich DTs in the translated question to be passed on to the Speaker LLM in a more easily understandable form, i.e. in the native language of the Speaker LLM. In this process, the domain-specific information captured by the Translator LLM\footnote{Such information-capturing capability is the strength of the Translator LLM, as this information cannot be captured independently by the Speaker LLM.} is explicitly conveyed to the Speaker LLM through the combination\footnote{Domain-specific terms (DTs) that are translated into the native language (LTs).} of DTs and LTs, further eliciting the relevant knowledge in the Speaker LLM and thereby achieving a collaborative joint enhancement effect.

\section{Experiments}
To explore the improvements that NatLan brings to knowledge elicitation, we selected question-answering (QA) as the evaluation task because it clearly indicates whether the relevant knowledge in the MLLMs has been correctly elicited.
Since the native language (dominant langauge) of nearly all mainstream MLLMs is English, we have selected \textit{English (en)} as the native language in this study. 

Considering the need to ensure diversity and distinctiveness among languages, we have selected five representative languages as our non-native target languages (hereafter referred to as the target language): \textit{Arabic (ar), Chinese (zh), French (fr), German (de), and Japanese (ja)}. Among these, French and German belong to the Indo-European language family, similar to English, while Arabic (Afro-Asiatic language family), Chinese (Sino-Tibetan language family), and Japanese (Japonic language family) are from different language families, showing significant differences from English.

\paragraph{Dataset.} 
Based on the aforementioned five target languages, we have selected the Multilingual MMLU (MMMLU) benchmark\footnote{\url{https://huggingface.co/datasets/openai/MMMLU}} of QA to assess the knowledge elicited from MLLMs. MMMLU consists of expert-translated versions of the MMLU benchmark \cite{hendrycks2021measuring} in different languages. Each language version comprises 14,079 multiple-choice questions from 57 disciplines. 

Additionally, we have also selected C-Eval Chinese benchmark \cite{huang2023c} to conduct further ablation studies and case studies, for the linguistic differences between Chinese and English are significant and both languages are sufficiently representative. C-Eval is also a multidisciplinary QA benchmark, containing 13,948 multiple-choice questions from 52 disciplines.

\paragraph{NatLan Setup.} 
In the proposed NatLan, we selected GPT-4o-mini as a universal translator to accomplish translations from the five target languages to the native language, for its comprehensive multilingual understanding capabilities. Additionally, to analyze the effects of the Translator LLM with varying capabilities on NatLan, we chose the Qwen MLLMs \cite{bai2023qwen} as Chinese-to-English translators, with 4B, 7B, and 14B parameters, for their leading Chinese comprehension capabilities. 

Furthermore, we selected five representative MLLMs with the capability to understand the five target languages to serve as Speakers, including models from the Phi \cite{abdin2024phi}, Gemma \cite{team2024gemma}, Mistral \cite{jiang2023mistral}, and Llama \cite{touvron2023llama2} series. For ease of joint deployment with the Translator LLMs, all these Speaker LLMs possess a moderate parameter scale, ranging from 3.8B to 7B.

\paragraph{Baselines.} Two top-notch related methods most relevant to the NatLan were selected as baselines: \textbf{(i) Self-Translation} \cite{zhang2023don, huang2023not, etxaniz2024multilingual}, which entails a single MLLM sequentially undertaking the translating and answering processes, serving both as the Translator LLM and the Speaker LLM.
\textbf{(ii) Google-MT} \cite{shi2022language}, which uses Google Neural Machine Translation system\footnote{Translation-only NMT model, unlike the general-purpose instruction-following MLLMs mentioned in this study.} (API) as the translator and MLLMs as the Speaker LLM. 

It is important to note that the requirement for Speaker LLMs to possess the five target languages comprehension abilities is crucial for conducting Self-Translation and direct evaluations on the target language, ensuring fair performance comparisons. More details are available in Appendix \ref{appendix:details}.

\subsection{Overall Performance Results} \label{sec:results}

\begin{table}[ht]
\centering
\resizebox{1\columnwidth}{!}{
\begin{tabular}{lccccc}
\hline
\textbf{Model} & \textbf{ar} & \textbf{zh} & \textbf{fr} & \textbf{de} & \textbf{ja} \\ 
\hline
Phi-3-mini (3.8B) & 33.66 & 43.04 & 57.49 & 55.06 & 41.38\\
\rowcolor{yellow!10}
+Self-Translation & 40.58 & 54.14 & 62.68 & 62.23 & 53.75\\
\rowcolor{yellow!10}
+Google-MT & 62.99 & 63.59 & 65.39 & 64.32 & 64.59\\
\rowcolor{green!10}
+NatLan & 64.18 & 64.23 & 65.55 & 65.58 & 65.00\\
+Human (Gold) & 68.10 & 68.10 & 68.10 & 68.10 & 68.10\\
\hline
Phi-3-small (7B) & 39.24 & 55.66 & 67.12 & 65.29 & 53.25\\
\rowcolor{yellow!10}
+Self-Translation & 53.04 & 64.10 & 66.22 & 68.10 & 63.76\\
\rowcolor{yellow!10}
+Google-MT & 68.35 & 69.56 & 71.46 & 69.14 & 70.32\\
\rowcolor{green!10}
+NatLan & 70.52 & 70.30 & 72.16 & 71.92 & 71.46\\
+Human (Gold) & 74.67 & 74.67 & 74.67 & 74.67 & 74.67\\
\hline
Gemma-1.1 (7B) & 39.72 & 47.70 & 50.35 & 49.36 & 45.11\\
\rowcolor{yellow!10}
+Self-Translation & 46.28 & 49.15 & 52.46 & 52.25 & 48.36\\
\rowcolor{yellow!10}
+Google-MT & 54.65 & 55.44 & 56.99 & 56.32 & 55.70\\
\rowcolor{green!10}
+NatLan & 56.10 & 56.03 & 56.84 & 56.72 & 56.55\\
+Human (Gold) & 58.12 & 58.12 & 58.12 & 58.12 & 58.12\\
\hline
Mistral-0.3 (7B) & 32.25 & 41.23 & 49.45 & 47.96 & 38.75\\
\rowcolor{yellow!10}
+Self-Translation & 39.99 & 46.04 & 52.32 & 52.26 & 46.13\\
\rowcolor{yellow!10}
+Google-MT & 54.49 & 55.46 & 57.14 & 55.59 & 55.85\\
\rowcolor{green!10}
+NatLan & 56.08 & 56.17 & 57.34 & 56.93 & 56.59\\
+Human (Gold) & 58.70 & 58.70 & 58.70 & 58.70 & 58.70\\
\hline
Llama-2 (7B) & 11.88 & 18.81 & 18.53 & 24.45 & 16.51\\
\rowcolor{red!10}
+Self-Translation & 10.78 & 15.16 & 20.05 & 17.42 & 12.69\\
\rowcolor{yellow!10}
+Google-MT & 32.55 & 31.80 & 33.22 & 32.74 & 29.92\\
\rowcolor{green!10}
+NatLan & 32.54 & 32.34 & 33.18 & 32.79 & 31.40\\
+Human (Gold) & 34.97 & 34.97 & 34.97 & 34.97 & 34.97\\
\hline
\end{tabular}
}
\caption{Performance on the MMMLU benchmark, measured by accuracy. Target languages are represented by their abbreviations. Red, yellow, and green indicate negative, suboptimal, and optimal enhancement, respectively. NatLan employs the universal translator, GPT-4o-mini. \textit{+Human (Gold)} represents the performance of the Speaker LLMs when answering on the original human-constructed English version MMLU benchmark.}
\label{tab: mmmlu-all}
\end{table}

As shown in Table \ref{tab: mmmlu-all}, we demonstrated the performance of the proposed NatLan, along with top-notch related methods, on the MMMLU benchmark. Overall, NatLan achieved accuracy surpassing all top-notch related methods. Furthermore, it closely approached the gold standard performance achieved when answering questions directly on the human expert-constructed English version.

More specifically, the performance of different methods exhibited a particular incremental pattern, i.e. \textit{Self-Translation < Google-MT < NatLan}. We confirmed in Section \S \ref{sec:DT_comp} that such performance differences are closely associated with the incremental retention of DTs in the questions translated by each method, where the retention follows the same order: \textit{Self-Translation < Google-MT < NatLan}. Moreover, virtually all translate-then-answer methods outperformed direct answers in the non-native language (reflecting the impact of LTs), validating the rationality of using LTs and DTs to interpret the translate-then-answer methods.

However, it is noteworthy that on a few MLLMs, particularly Llama-2 (7B), due to the limitations of the Speaker LLM's own instruction-following capabilities, the performance of Self-Translation was worse than answering directly in the non-native language. This degradation is caused by the propagation of errors during the translation phase, further emphasizing the necessity of incorporating additional models better suited for multilingual translation tasks to achieve more stable performance improvements from the translate-then-answer process. Further performance demonstrations on the C-Eval benchmark are available in Appendix \ref{appendix:c-eval}.

\begin{table*}[ht]
\centering
\resizebox{2\columnwidth}{!}{
\begin{tabular}{m{0.24\textwidth}m{0.3\textwidth}m{0.3\textwidth}m{0.16\textwidth}}
\hline
\textbf{Original Question} & \textbf{Google-MT Trans. Question} & \textbf{NatLan Trans. Question} & \textbf{Answers} \\ 
\hline
\begin{tabular}{@{}p{0.24\textwidth}@{}}
\begin{CJK*}{UTF8}{gbsn}
云南民俗中有 “女儿国” 和 “君子国” ，这 “两绝” 的形成与下列哪种因素有关$\_\_\_\_$。
\end{CJK*}\\
\begin{CJK*}{UTF8}{gbsn}
A. 生活水平低
\end{CJK*}\\
\begin{CJK*}{UTF8}{gbsn}
B. 文化素质差
\end{CJK*}\\
\begin{CJK*}{UTF8}{gbsn}
C. 交通闭塞
\end{CJK*}\\
\begin{CJK*}{UTF8}{gbsn}
D. 开发历史短
\end{CJK*}
\end{tabular}
& \begin{tabular}{@{}p{0.3\textwidth}@{}}
There are \textcolor{darkred}{``Daughter Country''} and \textcolor{darkred}{``Gentleman Country''} in Yunnan folklore. Which of the following factors is related to the formation of these ``two uniques''$\_\_\_\_$.\\
A. Low living standards\\
B. Poor \textcolor{darkred}{cultural quality}\\
C. \textcolor{darkred}{Impeded transportation}\\
D. Short development history
\end{tabular}
& \begin{tabular}{@{}p{0.3\textwidth}@{}}
The formation of \textcolor{darkgreen}{``the Kingdom of Women''} and \textcolor{darkgreen}{``the Kingdom of Gentlemen''} in Yunnan folklore is related to$\_\_\_\_$.\\
A. Low living standards\\
B. Poor \textcolor{darkgreen}{cultural literacy}\\
C. \textcolor{darkgreen}{Isolation due to poor transportation}\\
D. Short development history
\end{tabular}
& \begin{tabular}{@{}p{0.16\textwidth}@{}}
\\
Original: \textcolor{darkred}{B}\\\\
+Google-MT: \textcolor{darkred}{D}\\\\
+NatLan : \textcolor{darkgreen}{C}\\\\
True Label: \textcolor{darkgreen}{C}
\end{tabular} \\ 
\hline
\begin{tabular}{@{}p{0.24\textwidth}@{}}
\begin{CJK*}{UTF8}{gbsn}
单地址指令中为了完成两个数的算术运算，除地址码指明一个操作数外，另一个采用$\_\_\_\_$方式。
\end{CJK*}\\
\begin{CJK*}{UTF8}{gbsn}
A. 立即寻址
\end{CJK*}\\
\begin{CJK*}{UTF8}{gbsn}
B. 隐含寻址
\end{CJK*}\\
\begin{CJK*}{UTF8}{gbsn}
C. 间接寻址
\end{CJK*}\\
\begin{CJK*}{UTF8}{gbsn}
D. 基址寻址
\end{CJK*}
\end{tabular}
& \begin{tabular}{@{}p{0.3\textwidth}@{}}
In order to complete the arithmetic operation of two numbers in a single-address instruction, in addition to the address code indicating one operand, the other one uses $\_\_\_\_$ method.\\
A. Immediate addressing\\
B. Implicit addressing\\
C. Indirect addressing\\
D. Base addressing\\
\end{tabular}
& \begin{tabular}{@{}p{0.3\textwidth}@{}}
In a single-address instruction to perform arithmetic operations on two numbers, apart from the operand specified by the address code, the other one \textcolor{darkgreen}{is accessed} using the $\_\_\_\_$ method.\\
A. Immediate addressing\\
B. Implicit addressing\\
C. Indirect addressing\\
D. Base addressing\\
\end{tabular}
& \begin{tabular}{@{}p{0.16\textwidth}@{}}
\\
Original: \textcolor{darkred}{C}\\\\
+Google-MT: \textcolor{darkred}{C}\\\\
+NatLan : \textcolor{darkgreen}{B}\\\\
True Label : \textcolor{darkgreen}{B}
\end{tabular} \\ 
\hline
\end{tabular}
}
\caption{Chinese-to-English translation cases in C-Eval. More cases and details are available in Appendix \ref{appendix:cases}.}
\label{tab:case-study}
\end{table*}

\subsection{NatLan Retains More Domain Triggers} \label{sec:DT_comp}

To effectively assess the superiority of the proposed NatLan over other translate-then-answer baselines in DT retention, we developed a pairwise comparison evaluation method. Under the supervision of GPT-4o-mini, this evaluation method examines the advantage ratios of different approaches in DT retention across all samples in the MMMLU benchmark. Specifically, through few-shot prompting, GPT-4o-mini selects the translation with better DT retention from two given options, determining which method is superior for that sample.

\begin{figure}[ht] 
\centering
\includegraphics[width=1\linewidth]{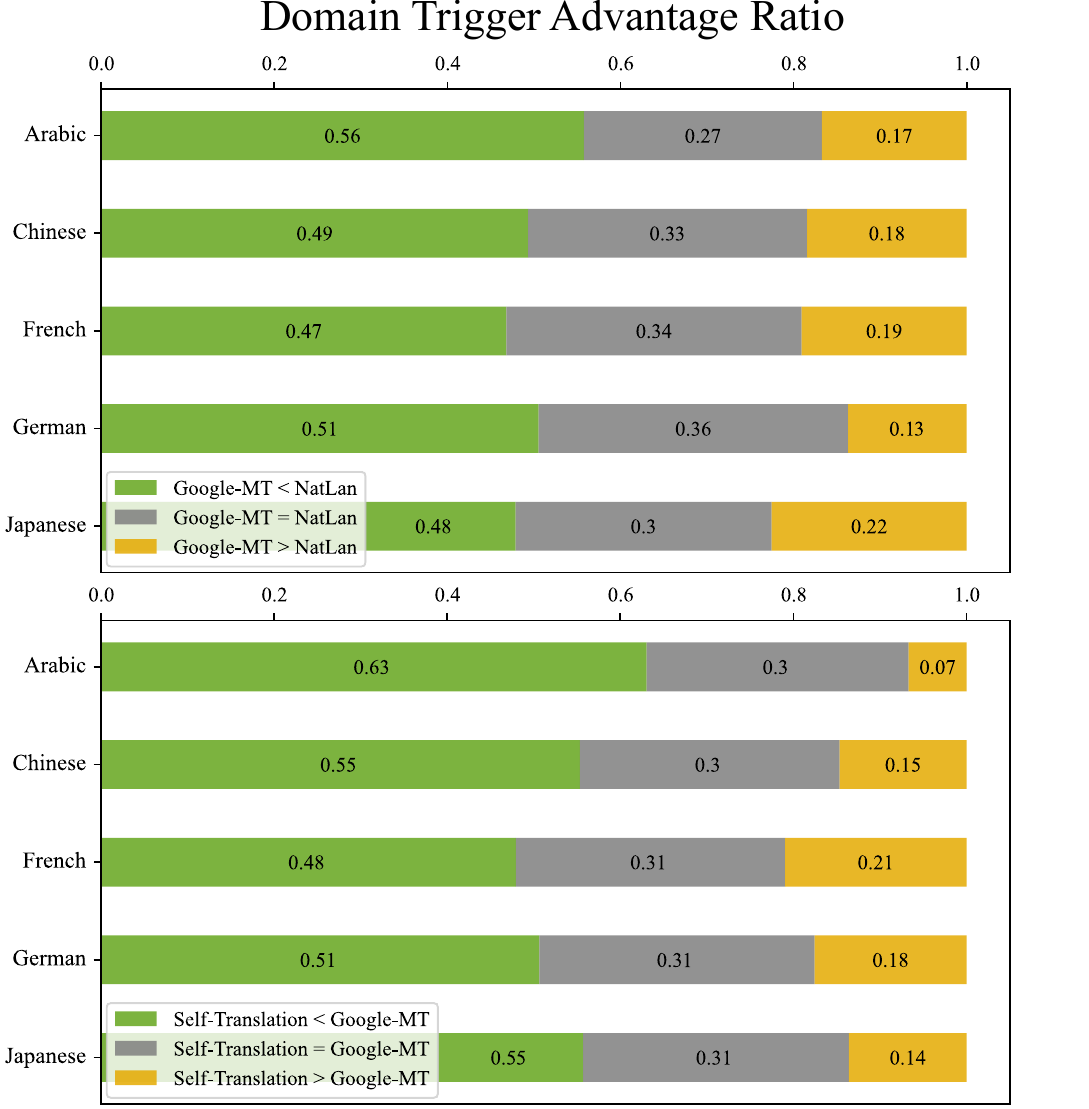}
\caption{Pairwise compared DT advantage ratios on the MMMLU benchmark under GPT-4o-mini supervision. The gray bars indicate that the retention of DTs in the questions translated by the two methods is nearly equivalent. More details are available in Appendix \ref{appendix:details}.}
\label{fig:DT_comp}
\end{figure}

As depicted in Figure \ref{fig:DT_comp}, the results of the pairwise comparison clearly demonstrate the same incremental trend reported in \S \ref{sec:results}, i.e., in terms of DT retention, \textit{Self-Translation} (Phi-3-small (7B)) < \textit{Google-MT < NatLan}. In other words, the DT retention achieved during the translation process is \textbf{positively} correlated with the accuracy of the answers based on the translated questions. This confirms the validity of using DTs to interpret the mechanisms of the translate-then-answer methods and emphasizes the effectiveness of NatLan's improvements aimed at enhancing DT retention.

\subsection{Two-Aspect Domain Trigger Retention} \label{sec:case_study}

We conducted a comprehensive analysis to determine how NatLan retains DTs from the original questions in translated cases, summarizing the findings into the following two principal aspects:

\paragraph{(i) Accurate descriptions of domain-specific entities.}
Precisely translating descriptions specific to the domain of the entities, avoiding overly generic or literal translations.   
As shown in the first row of Table \ref{tab:case-study}, Google-MT produces a literal translation such as \underline{``Daughter Country''} without considering the folkloric context of the Chinese expression, whereas NatLan accurately uses \underline{``the Kingdom of Women''}.   Similarly, NatLan uses \underline{``literacy''} instead of \underline{``quality''}, etc., rendering the descriptions of the translated entities semantically more accurate in the folkloric context of this case.

\paragraph{(ii) Explicit descriptions of relationships between domain-specific entities.}
Although it is possible to approximately infer the implicit relationships between domain-specific entities through some deliberation, this clearly increases cognitive load and does not guarantee the accuracy of such inferences. Explicitly describing domain-specific operations between entities can effectively avoid ambiguity from implicit descriptions. As shown in the first row of Table \ref{tab:case-study}, NatLan explicitly renders the \underline{``is accessed''} operation between the operand and the addressing method, conveying a more complete translation of the domain-specific relationships implied in the original question.

\paragraph{}
It should be noted that since Self-Translation is inferior to Google-MT, to fairly demonstrate the advantages of NatLan in DT retention, we have only compared NatLan with the Google-MT.

\begin{figure*}[ht] 
\centering
\includegraphics[width=1\linewidth]{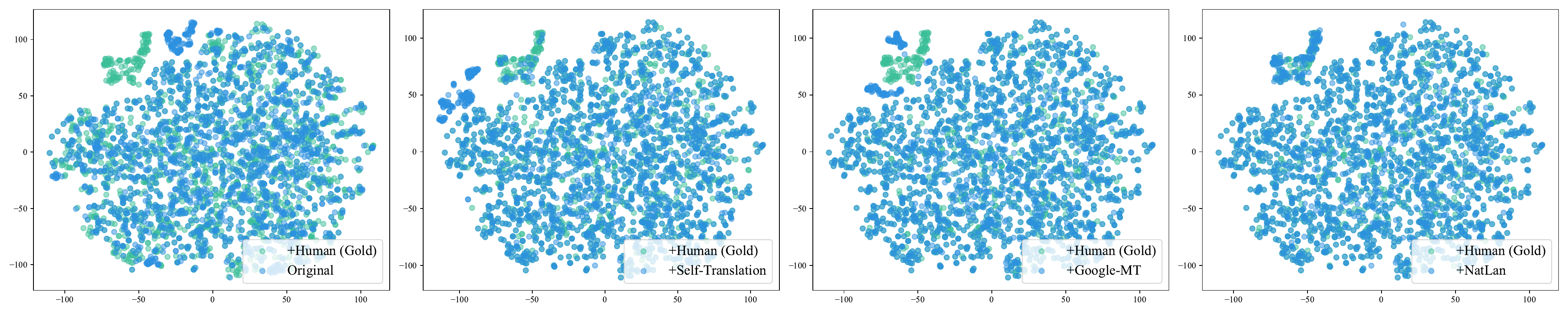}
\caption{The visualized knowledge activation distributions on the French version of the MMMLU benchmark, with the Speaker LLM: Phi-3-small (7B). The greater the overlap with the green dots (human gold standard), the more accurate the knowledge activation is considered. More cases are available in Appendix \ref{appendix:act}.}
\label{act_analysis_1}
\end{figure*}

\subsection{Knowledge Activation Rectification} \label{sec:act_analysis}
We further analyzed how increased retention of DTs in translation results impacts knowledge elicitation during the answering process at a more fine-grained activation level.
In our QA task setup, since the Speaker LLMs only need to generate the answer options, the last hidden state for predicting the first token reflects the internal knowledge activation pattern used for answer generation. Therefore, we extracted it for knowledge activation analysis.
Additionally, on the MMMLU benchmark, we used the knowledge activation generated by Speaker LLMs when answering on the human-constructed English version MMLU benchmark as the gold knowledge activation standard\footnote{This also corresponds to the optimal performance that Speaker LLMs can achieve on this benchmark.} for evaluation.

We initially visualized the knowledge activation distribution of Speaker LLMs using t-SNE \cite{van2008visualizing}, as depicted in Figure \ref{act_analysis_1}. The blue dots represent the knowledge activation distributions obtained from various methods on the MMMLU. From left to right, the methods used are: answering directly on non-native language questions (Original), Self-Translation, Google-MT, and our proposed NatLan. Qualitatively, compared to the knowledge activation distribution obtained when answering on the human gold standard (Green), NatLan shows a higher degree of overlap.

\begin{table}[ht]
\centering
\resizebox{1\columnwidth}{!}{
\begin{tabular}{lrrrrr}
\hline
\textbf{Method} & \textbf{ar} & \textbf{zh} & \textbf{fr} & \textbf{de} & \textbf{ja}\\ 
\hline
Original & 111.40 & 69.54 & 21.50 & 36.72 & 87.43\\
\rowcolor{yellow!10}
+Self-Translation & 35.31 & 20.86 & 15.60 & 15.76 & 22.31\\
\rowcolor{yellow!10}
+Google-MT & 13.24 & 12.84 & 9.06 & 10.07 & 13.88\\
\rowcolor{green!10}
+NatLan & 9.84 & 11.24 & 7.46 & 8.38 & 11.28\\
\hline
\end{tabular}
}
\caption{Average Euclidean distances between the knowledge activation (extracted from Phi-3-small (7B)) obtained using various methods and that obtained using the human gold standard. A smaller distance indicates a closer approximation to the human gold standard.}
\label{tab: act_dist}
\end{table}


Moreover, on the MMMLU benchmark, we measured the average Euclidean distance between the activation vectors obtained through various methods and those obtained when answering using the human gold standard. As shown in Table \ref{tab: act_dist}, NatLan achieved the closest approximation to the human gold standard in terms of knowledge activation. Interestingly, the performance evaluation of activation distance followed the same incremental trend previously mentioned, namely: \textit{Self-Translation < Google-MT < NatLan}. This uniformly confirms that more complete DT retention after translation can rectify the activation in Speaker LLMs when answering questions, eliciting relevant knowledge.

\begin{figure}[ht] 
\centering
\includegraphics[width=1\linewidth]{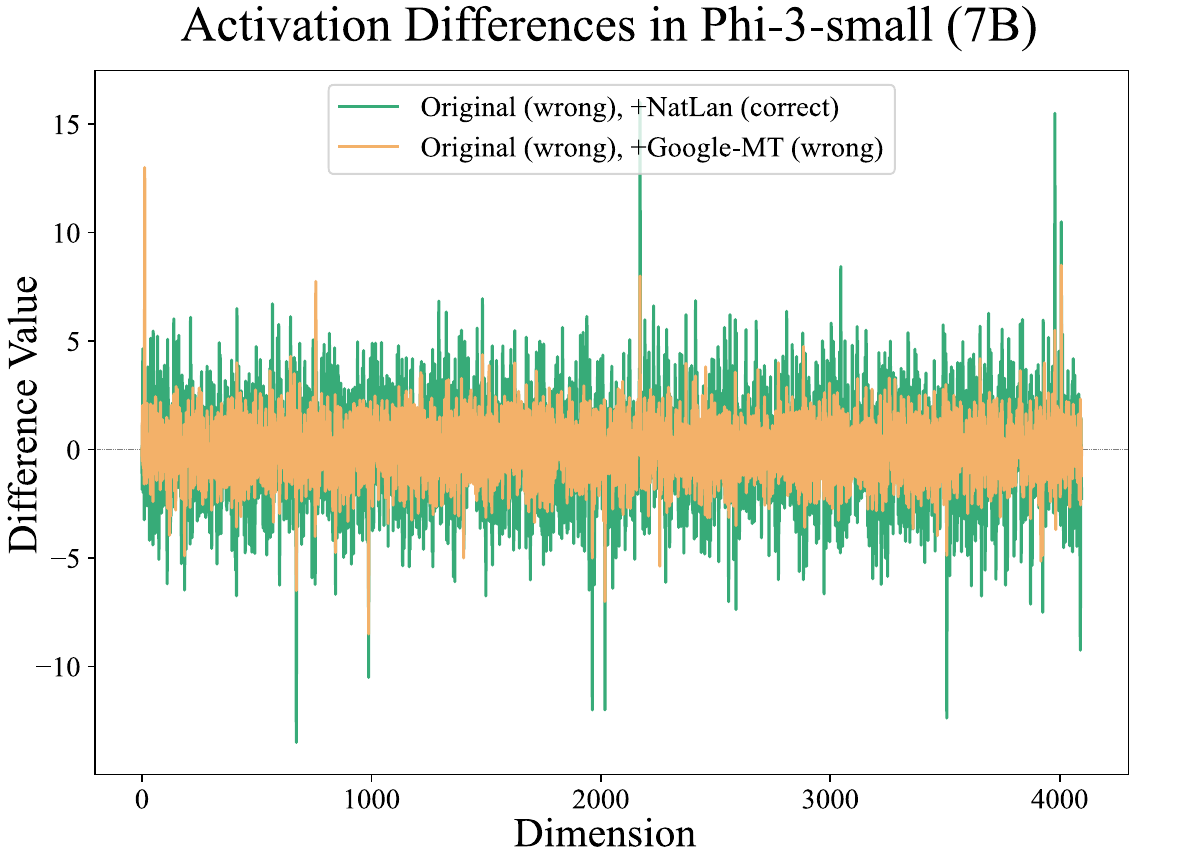}
\caption{Activation differences between different methods for the same questions. Contents in parentheses indicate the correctness of the Speaker LLMs' responses.}
\label{act_analysis_diff}
\end{figure}

Furthermore, from Figure \ref{act_analysis_diff}, we can observe that Google-MT, the best-performing related methods, also provides some degree of knowledge activation rectification. However, due to its inferior DT retention compared to our proposed NatLan, the rectification effect brought by Google-MT is insufficient to correct the answers. In contrast, NatLan, through more complete DT retention, delivers a higher level of rectification effect, thereby successfully correcting originally incorrect answers.


\subsection{Impact of Translator LLMs' Semantic Capabilities on NatLan} \label{sec:ablation_1}

\begin{table}[ht]
\centering
\resizebox{1\columnwidth}{!}{
\begin{tabular}{lccc}
\hline
\textbf{Model} & \textbf{Lang.} & \textbf{Avg.} & \textbf{Avg. (Hard)} \\ 
\hline
\multicolumn{4}{c}{\textit{Translator LLMs}}\\
\hline
Qwen-1.5 (4B) & zh & 60.1 & 42.3\\
Qwen-2 (7B) & zh & 78.9 & 56.7\\
Qwen-1.5 (14B) & zh & 74.9 & 58.9\\
\hline
\multicolumn{4}{c}{\textit{Speaker LLMs}}\\
\hline
Phi-3-mini (3.8B) & zh & 41.2 & 36.3\\
+NatLan Qwen-1.5 (4B) & en & 48.1 & 37.9\\
+NatLan Qwen-2 (7B) & en & 50.8 & 39.9\\
\rowcolor{green!10}
+NatLan Qwen-1.5 (14B) & en & 51.3 & 41.3\\
\hline
Phi-3-small (7B) & zh & 49.0 & 41.6\\
+NatLan Qwen-1.5 (4B) & en & 52.7 & 41.9\\
+NatLan Qwen-2 (7B) & en & 56.0 & 43.5\\
\rowcolor{green!10}
+NatLan Qwen-1.5 (14B) & en & 55.9 & 44.7\\
\hline
Gemma-1.1 (7B) & zh & 44.4 & 36.3\\
+NatLan Qwen-1.5 (4B) & en & 45.0 & 38.2\\
\rowcolor{green!10}
+NatLan Qwen-2 (7B) & en & 47.7 & 38.6\\
+NatLan Qwen-1.5 (14B) & en & 47.6 & 38.0\\
\hline
Mistral-0.3 (7B) & zh & 42.8 & 32.6\\
+NatLan Qwen-1.5 (4B) & en & 45.6 & 33.6\\
\rowcolor{green!10}
+NatLan Qwen-2 (7B) & en & 48.4 & 35.3\\
+NatLan Qwen-1.5 (14B) & en & 47.8 & 35.5\\
\hline
Llama-2 (7B) & zh & 21.3 & 14.7\\
+NatLan Qwen-1.5 (4B) & en & 25.6 & 18.7\\
+NatLan Qwen-2 (7B) & en & 25.2 & 17.3\\
\rowcolor{green!10}
+NatLan Qwen-1.5 (14B) & en & 27.6 & 18.6\\
\hline
\end{tabular}
}
\caption{Performance comparison of NatLan using different Translator LLMs on the C-Eval test sets.}
\label{tab: ablation_natlan}
\end{table}

\begin{figure*}[ht] 
\centering
\includegraphics[width=1\linewidth]{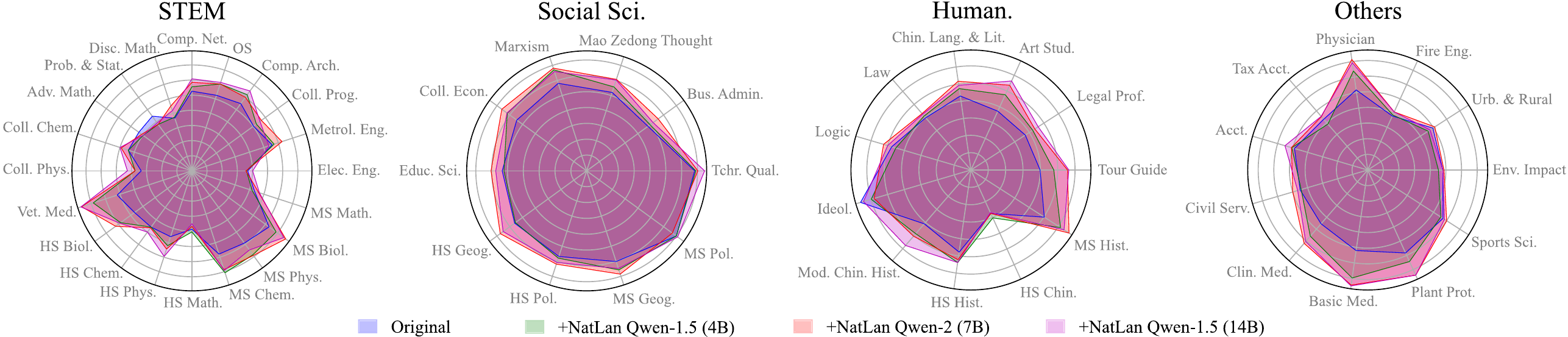}
\caption{Performance of NatLan across 52 disciplines on the C-Eval test sets under four configurations, with Phi-3-small (7B) as the Speaker LLM. More details are available in Appendix \ref{appendix:ablation}.}
\label{phi3_radar}
\end{figure*}

To analyze the correlation between the semantic capabilities of the Translator LLM and the extent of improvements it can provide within NatLan, we employed three Qwen MLLMs with varying Chinese semantic understanding abilities as Chinese-to-English Translator LLMs and conducted a comparative analysis of performance gains on the C-Eval Chinese benchmark. As shown in Table \ref{tab: ablation_natlan}, Qwen-2 (7B) and Qwen-1.5 (14B) exhibit comparable semantic capabilities, each with their own strengths, while Qwen-1.5 (4B) has relatively weaker semantic capabilities in comparison. Furthermore, when they serve as Translator LLMs, the relative strengths and weaknesses of their semantic capabilities are generally reflected in the varying degrees of knowledge elicitation from the Speaker LLMs.

Specifically, NatLan Qwen-2 (7B) and NatLan Qwen-1.5 (14B) generally provide comparable performance improvements. The former tends to perform better in terms of average accuracy across most models, while the latter excels in average accuracy at the hard level, aligning with their respective strengths. This confirms the pivotal role of the semantic capabilities of Translator LLMs in the effectiveness of NatLan. Intuitively, this impact applies to all translate-then-answer methods.

\subsection{Impact of Domain-Language Characteristics on NatLan} \label{sec:ablation_2}
More comprehensively, we explored the characteristics that exist between specific domains and languages and their impact on NatLan. As shown in Figure \ref{phi3_radar}, although using NatLan to translate questions from a non-native language (Chinese) to the native language (English) greatly enhances answer accuracy across most disciplines, due to the richer knowledge available in the native language across most domains, there are exceptions. In a few disciplines, translating languages using NatLan does not result in performance gains. We categorized such disciplines with specific domain-language characteristics into the following two types:

\paragraph{(i) Language-Insensitive Disciplines:}
 Disciplines like Probability and Statistics (Prob. \& Stat.), which mostly rely on understanding mathematical formulas to answer questions. These mathematical formulas are consistent across languages. In such cases, adding a translation process can introduce potential errors, such as the loss of content in mathematical formulas during translation, impacting the correct understanding of the questions. 

\paragraph{(ii) Language-Knowledge Bound Disciplines:}
 Disciplines like Ideological and Moral Cultivation (Ideol.), which are closely tied to specific languages due to cultural and other differences between countries. In these disciplines, training materials in different languages may contain distinctly different relevant knowledge about similar topics. Uniformly translating such questions into the native language (English) could lead to confusion and conflicts in the relevant knowledge, impacting the correct elicitation of knowledge.

\paragraph{}
It should be noted that while there is a possibility that applying NatLan in the aforementioned two types of disciplines may not offer benefits, this is not always the case. The actual occurrence still largely depends on the capabilities of the Translator LLMs and Speaker LLMs involved.

\section{Conclusion}

MLLMs often fail to answer questions posed in non-dominant languages, which they can successfully address when presented in their dominant languages. Although existing translate-then-answer methods can mitigate this issue, the mechanisms behind their effectiveness remain unclear. To clarify, we employ human cognitive features: Language Triggers (LTs) and Domain Triggers (DTs) to interpret the mechanisms behind existing translate-then-answer methods, highlighting issues of DT insufficiency in existing approaches and proposing NatLan as a remediation. Our proposed NatLan achieves up to a \textbf{31.28\%} average accuracy improvement across five non-native language QA benchmarks and provides comparable or greater retention of DTs in up to \textbf{87\%} of cases compared to Google-MT, advancing the understanding of translate-then-answer process through the lens of LTs and DTs.

\section*{Limitations}

The Speaker LLMs selected for this study all use English as their dominant language (native language). Although we aimed to assess MLLMs with various native languages, the vast majority of existing MLLMs primarily utilize English as their native language. Even if some MLLMs demonstrate stronger capabilities in other languages, such as the enhanced proficiency in Chinese of Qwen MLLMs, they still cannot outperform the performance under English prompting. In other words, English corpora consistently dominate their training data. In our preliminary investigation, Qwen-2 (7B) achieved only \textbf{47.70\%} accuracy on the human expert-constructed Chinese version of MMLU, significantly lower than the \textbf{58.19\%} accuracy attained on the English version of MMLU. Therefore, we encourage future research to develop MLLMs with different native languages other than English, or investigate other advanced language transfer techniques. Such explorations could have a profound impact on the development of applications for low-resource languages.

Furthermore, although NatLan significantly enhances the performance of MLLMs, the potential improvements attributable to NatLan are inherently limited by the capabilities of the Translator LLMs and particularly the Speaker LLMs, where the primary bottlenecks tend to occur. Moreover, as observed in the analysis from \S \ref{sec:ablation_2}, for a minority of disciplines, NatLan fails to enhance performance. In addition to translation errors produced by Translator LLMs, another significant factor is that some knowledge is closely tied to specific languages, such as in the Ideology and Moral Cultivation discipline. Employing the native language to address these types of issues may not yield benefits and could instead prevent the successful recall of relevant knowledge. Therefore, we encourage future work to explore the scope of knowledge covered by various languages in MLLMs, aiming to achieve an adaptive and dynamic language switching during question-answering, specifically switching to the language that best encompasses the required knowledge for optimal knowledge elicitation.

\section*{Ethical Considerations}

LLMs are prone to generating incorrect and potentially biased information. This issue becomes especially significant when LLMs are tasked with responding to sensitive questions. While NatLan enhances the performance of LLMs, it does not eliminate the issue of producing biased or incorrect statements. In light of some potential issues, this study advocates for usage under research purposes. Cautious deployment is advisable when integrating such systems into user-facing applications.

All the datasets and models used in this study are publicly available with permissible licenses. The C-Eval Chinese benchmark have CC-BY-NC-SA-4.0 License \footnote{\url{https://spdx.org/licenses/CC-BY-NC-SA-4.0}}, the MMMLU benchmark and Phi-3-* models have MIT License \footnote{\url{https://choosealicense.com/licenses/mit}}, Qwen-1.5-* models have Tongyi-Qianwen-Research License \footnote{\url{https://huggingface.co/Qwen/Qwen1.5-14B-Chat/blob/main/LICENSE}}, Qwen-2-* and Mistral-0.3-* models have Apache-2.0 License \footnote{\url{https://choosealicense.com/licenses/apache-2.0}}, Llama-2-* models have Llama 2 Community License \footnote{\url{https://huggingface.co/meta-llama/Llama-2-7b-chat-hf/blob/main/LICENSE.txt}} and Gemma-1.1-* models have Gemma Terms of Use \footnote{\url{https://ai.google.dev/gemma/terms}}.

\bibliography{acl_latex}

\begin{thebibliography}{32}
\expandafter\ifx\csname natexlab\endcsname\relax\def\natexlab#1{#1}\fi

\bibitem[{Abdin et~al.(2024)Abdin, Jacobs, Awan, Aneja, Awadallah, Awadalla, Bach, Bahree, Bakhtiari, Behl et~al.}]{abdin2024phi}
Marah Abdin, Sam~Ade Jacobs, Ammar~Ahmad Awan, Jyoti Aneja, Ahmed Awadallah, Hany Awadalla, Nguyen Bach, Amit Bahree, Arash Bakhtiari, Harkirat Behl, et~al. 2024.
\newblock \href {https://arxiv.org/abs/2404.14219} {Phi-3 technical report: A highly capable language model locally on your phone}.
\newblock \emph{arXiv preprint arXiv:2404.14219}.

\bibitem[{Acheampong et~al.(2016)Acheampong, Pan, Zhou, and Li}]{acheampong2016answer}
Kingsley~N Acheampong, Zhen-Hao Pan, Er-Qiang Zhou, and Xiao-Yu Li. 2016.
\newblock \href {https://ieeexplore.ieee.org/abstract/document/8079800} {Answer triggering of factoid questions: A cognitive approach}.
\newblock In \emph{2016 13th International Computer Conference on Wavelet Active Media Technology and Information Processing (ICCWAMTIP)}, pages 33--37. IEEE.

\bibitem[{Achiam et~al.(2023)Achiam, Adler, Agarwal, Ahmad, Akkaya, Aleman, Almeida, Altenschmidt, Altman, Anadkat et~al.}]{achiam2023gpt}
Josh Achiam, Steven Adler, Sandhini Agarwal, Lama Ahmad, Ilge Akkaya, Florencia~Leoni Aleman, Diogo Almeida, Janko Altenschmidt, Sam Altman, Shyamal Anadkat, et~al. 2023.
\newblock \href {https://arxiv.org/abs/2303.08774} {Gpt-4 technical report}.
\newblock \emph{arXiv preprint arXiv:2303.08774}.

\bibitem[{Ai et~al.(2023)Ai, He, Yu, and Wang}]{ai2023tecs}
Yiming Ai, Zhiwei He, Kai Yu, and Rui Wang. 2023.
\newblock \href {https://aclanthology.org/2023.acl-short.164/} {Tecs: A dataset and benchmark for tense consistency of machine translation}.
\newblock In \emph{Proceedings of the 61st Annual Meeting of the Association for Computational Linguistics (Volume 2: Short Papers)}, pages 1930--1941.

\bibitem[{Bai et~al.(2023)Bai, Bai, Chu, Cui, Dang, Deng, Fan, Ge, Han, Huang et~al.}]{bai2023qwen}
Jinze Bai, Shuai Bai, Yunfei Chu, Zeyu Cui, Kai Dang, Xiaodong Deng, Yang Fan, Wenbin Ge, Yu~Han, Fei Huang, et~al. 2023.
\newblock \href {https://arxiv.org/abs/2309.16609} {Qwen technical report}.
\newblock \emph{arXiv preprint arXiv:2309.16609}.

\bibitem[{Baker and Levin(2015)}]{baker2015role}
Lewis~J Baker and Daniel~T Levin. 2015.
\newblock \href {https://www.sciencedirect.com/science/article/abs/pii/S0010027714002509} {The role of relational triggers in event perception}.
\newblock \emph{Cognition}, 136:14--29.

\bibitem[{Brown et~al.(2020)Brown, Mann, Ryder, Subbiah, Kaplan, Dhariwal, Neelakantan, Shyam, Sastry, Askell et~al.}]{brown2020language}
Tom Brown, Benjamin Mann, Nick Ryder, Melanie Subbiah, Jared~D Kaplan, Prafulla Dhariwal, Arvind Neelakantan, Pranav Shyam, Girish Sastry, Amanda Askell, et~al. 2020.
\newblock \href {https://proceedings.neurips.cc/paper/2020/hash/1457c0d6bfcb4967418bfb8ac142f64a-Abstract.html} {Language models are few-shot learners}.
\newblock \emph{Advances in neural information processing systems}, 33:1877--1901.

\bibitem[{Del~Maschio et~al.(2022)Del~Maschio, Crespi, Peressotti, Abutalebi, and Sulpizio}]{del2022decision}
Nicola Del~Maschio, Federico Crespi, Francesca Peressotti, Jubin Abutalebi, and Simone Sulpizio. 2022.
\newblock \href {https://www.cambridge.org/core/journals/bilingualism-language-and-cognition/article/decisionmaking-depends-on-language-a-metaanalysis-of-the-foreign-language-effect/357CDDF948DB7FC3C101534F1EA3442A} {Decision-making depends on language: A meta-analysis of the foreign language effect}.
\newblock \emph{Bilingualism: Language and Cognition}, 25(4):617--630.

\bibitem[{Dong et~al.(2024)Dong, Jiang, Jin, and Li}]{dong2024self}
Yihong Dong, Xue Jiang, Zhi Jin, and Ge~Li. 2024.
\newblock \href {https://doi.org/10.1145/3672459} {Self-collaboration code generation via chatgpt}.
\newblock \emph{ACM Trans. Softw. Eng. Methodol.}

\bibitem[{Etxaniz et~al.(2024)Etxaniz, Azkune, Soroa, Lacalle, and Artetxe}]{etxaniz2024multilingual}
Julen Etxaniz, Gorka Azkune, Aitor Soroa, Oier Lacalle, and Mikel Artetxe. 2024.
\newblock \href {https://aclanthology.org/2024.naacl-short.46/} {Do multilingual language models think better in english?}
\newblock In \emph{Proceedings of the 2024 Conference of the North American Chapter of the Association for Computational Linguistics: Human Language Technologies (Volume 2: Short Papers)}, pages 550--564.

\bibitem[{Gao et~al.(2023)Gao, Hua, Armada-da Silva, Zhang, Li, Chen, Wang, Du, and Yuan}]{gao2023shared}
Fei Gao, Lin Hua, Paulo Armada-da Silva, Juan Zhang, Defeng Li, Zhiyi Chen, Chengwen Wang, Meng Du, and Zhen Yuan. 2023.
\newblock \href {https://www.nature.com/articles/s41539-023-00184-9} {Shared and distinct neural correlates of first and second language morphological processing in bilingual brain}.
\newblock \emph{npj Science of Learning}, 8(1):33.

\bibitem[{Guo et~al.(2024)Guo, Ren, Hu, Li, Zhang, Zhang, and Huang}]{guo2024teaching}
Ping Guo, Yubing Ren, Yue Hu, Yunpeng Li, Jiarui Zhang, Xingsheng Zhang, and He-Yan Huang. 2024.
\newblock \href {https://aclanthology.org/2024.lrec-main.1362/} {Teaching large language models to translate on low-resource languages with textbook prompting}.
\newblock In \emph{Proceedings of the 2024 Joint International Conference on Computational Linguistics, Language Resources and Evaluation (LREC-COLING 2024)}, pages 15685--15697.

\bibitem[{Hendrycks et~al.(2021)Hendrycks, Burns, Basart, Zou, Mazeika, Song, and Steinhardt}]{hendrycks2021measuring}
Dan Hendrycks, Collin Burns, Steven Basart, Andy Zou, Mantas Mazeika, Dawn Song, and Jacob Steinhardt. 2021.
\newblock \href {https://openreview.net/forum?id=d7KBjmI3GmQ} {Measuring massive multitask language understanding}.
\newblock In \emph{International Conference on Learning Representations}.

\bibitem[{Huang et~al.(2023{\natexlab{a}})Huang, Tang, Zhang, Zhao, Song, Xia, and Wei}]{huang2023not}
Haoyang Huang, Tianyi Tang, Dongdong Zhang, Wayne~Xin Zhao, Ting Song, Yan Xia, and Furu Wei. 2023{\natexlab{a}}.
\newblock \href {https://aclanthology.org/2023.findings-emnlp.826/} {Not all languages are created equal in llms: Improving multilingual capability by cross-lingual-thought prompting}.
\newblock In \emph{Findings of the Association for Computational Linguistics: EMNLP 2023}, pages 12365--12394.

\bibitem[{Huang et~al.(2023{\natexlab{b}})Huang, Bai, Zhu, Zhang, Zhang, Su, Liu, Lv, Lei, Fu et~al.}]{huang2023c}
Yuzhen Huang, Yuzhuo Bai, Zhihao Zhu, Junlei Zhang, Jinghan Zhang, Tangjun Su, Junteng Liu, Chuancheng Lv, Jiayi Lei, Yao Fu, et~al. 2023{\natexlab{b}}.
\newblock \href {https://proceedings.neurips.cc/paper_files/paper/2023/hash/c6ec1844bec96d6d32ae95ae694e23d8-Abstract-Datasets_and_Benchmarks.html} {C-eval: a multi-level multi-discipline chinese evaluation suite for foundation models}.
\newblock In \emph{Proceedings of the 37th International Conference on Neural Information Processing Systems}, pages 62991--63010.

\bibitem[{Jiang et~al.(2023)Jiang, Sablayrolles, Mensch, Bamford, Chaplot, Casas, Bressand, Lengyel, Lample, Saulnier et~al.}]{jiang2023mistral}
Albert~Q Jiang, Alexandre Sablayrolles, Arthur Mensch, Chris Bamford, Devendra~Singh Chaplot, Diego de~las Casas, Florian Bressand, Gianna Lengyel, Guillaume Lample, Lucile Saulnier, et~al. 2023.
\newblock \href {https://arxiv.org/abs/2310.06825} {Mistral 7b}.
\newblock \emph{arXiv preprint arXiv:2310.06825}.

\bibitem[{Kang et~al.(2024)Kang, Blevins, and Zettlemoyer}]{kang2024translate}
Haoqiang Kang, Terra Blevins, and Luke Zettlemoyer. 2024.
\newblock \href {https://aclanthology.org/2024.eacl-long.94/} {Translate to disambiguate: Zero-shot multilingual word sense disambiguation with pretrained language models}.
\newblock In \emph{Proceedings of the 18th Conference of the European Chapter of the Association for Computational Linguistics (Volume 1: Long Papers)}, pages 1562--1575.

\bibitem[{Liu et~al.(2022)Liu, Chen, and Xu}]{liu2022saliency}
Jian Liu, Yufeng Chen, and Jinan Xu. 2022.
\newblock \href {https://aclanthology.org/2022.acl-long.313/} {Saliency as evidence: Event detection with trigger saliency attribution}.
\newblock In \emph{Proceedings of the 60th Annual Meeting of the Association for Computational Linguistics}, pages 4573--4585.

\bibitem[{Schulhoff et~al.(2024)Schulhoff, Ilie, Balepur, Kahadze, Liu, Si, Li, Gupta, Han, Schulhoff et~al.}]{schulhoff2024prompt}
Sander Schulhoff, Michael Ilie, Nishant Balepur, Konstantine Kahadze, Amanda Liu, Chenglei Si, Yinheng Li, Aayush Gupta, HyoJung Han, Sevien Schulhoff, et~al. 2024.
\newblock \href {https://arxiv.org/abs/2406.06608} {The prompt report: A systematic survey of prompting techniques}.
\newblock \emph{arXiv preprint arXiv:2406.06608}.

\bibitem[{Shi et~al.(2022)Shi, Suzgun, Freitag, Wang, Srivats, Vosoughi, Chung, Tay, Ruder, Zhou et~al.}]{shi2022language}
Freda Shi, Mirac Suzgun, Markus Freitag, Xuezhi Wang, Suraj Srivats, Soroush Vosoughi, Hyung~Won Chung, Yi~Tay, Sebastian Ruder, Denny Zhou, et~al. 2022.
\newblock \href {https://openreview.net/forum?id=fR3wGCk-IXp} {Language models are multilingual chain-of-thought reasoners}.
\newblock In \emph{The Eleventh International Conference on Learning Representations}.

\bibitem[{Talebirad and Nadiri(2023)}]{talebirad2023multi}
Yashar Talebirad and Amirhossein Nadiri. 2023.
\newblock \href {https://arxiv.org/abs/2306.03314} {Multi-agent collaboration: Harnessing the power of intelligent llm agents}.
\newblock \emph{arXiv preprint arXiv:2306.03314}.

\bibitem[{Team et~al.(2024)Team, Mesnard, Hardin, Dadashi, Bhupatiraju, Pathak, Sifre, Rivi{\`e}re, Kale, Love et~al.}]{team2024gemma}
Gemma Team, Thomas Mesnard, Cassidy Hardin, Robert Dadashi, Surya Bhupatiraju, Shreya Pathak, Laurent Sifre, Morgane Rivi{\`e}re, Mihir~Sanjay Kale, Juliette Love, et~al. 2024.
\newblock \href {https://arxiv.org/abs/2403.08295} {Gemma: Open models based on gemini research and technology}.
\newblock \emph{arXiv preprint arXiv:2403.08295}.

\bibitem[{Tong et~al.(2020)Tong, Xu, Wang, Cao, Hou, Li, and Xie}]{tong2020improving}
Meihan Tong, Bin Xu, Shuai Wang, Yixin Cao, Lei Hou, Juanzi Li, and Jun Xie. 2020.
\newblock \href {https://aclanthology.org/2020.acl-main.522/} {Improving event detection via open-domain trigger knowledge}.
\newblock In \emph{Proceedings of the 58th Annual Meeting of the Association for Computational Linguistics}, pages 5887--5897.

\bibitem[{Touvron et~al.(2023{\natexlab{a}})Touvron, Lavril, Izacard, Martinet, Lachaux, Lacroix, Rozi{\`e}re, Goyal, Hambro, Azhar et~al.}]{touvron2023llama}
Hugo Touvron, Thibaut Lavril, Gautier Izacard, Xavier Martinet, Marie-Anne Lachaux, Timoth{\'e}e Lacroix, Baptiste Rozi{\`e}re, Naman Goyal, Eric Hambro, Faisal Azhar, et~al. 2023{\natexlab{a}}.
\newblock \href {https://arxiv.org/abs/2302.13971} {Llama: Open and efficient foundation language models}.
\newblock \emph{arXiv preprint arXiv:2302.13971}.

\bibitem[{Touvron et~al.(2023{\natexlab{b}})Touvron, Martin, Stone, Albert, Almahairi, Babaei, Bashlykov, Batra, Bhargava, Bhosale et~al.}]{touvron2023llama2}
Hugo Touvron, Louis Martin, Kevin Stone, Peter Albert, Amjad Almahairi, Yasmine Babaei, Nikolay Bashlykov, Soumya Batra, Prajjwal Bhargava, Shruti Bhosale, et~al. 2023{\natexlab{b}}.
\newblock \href {https://arxiv.org/abs/2307.09288} {Llama 2: Open foundation and fine-tuned chat models}.
\newblock \emph{arXiv preprint arXiv:2307.09288}.

\bibitem[{Tu et~al.(2017)Tu, Liu, Shang, Liu, and Li}]{tu2017neural}
Zhaopeng Tu, Yang Liu, Lifeng Shang, Xiaohua Liu, and Hang Li. 2017.
\newblock \href {https://ojs.aaai.org/index.php/AAAI/article/view/10950} {Neural machine translation with reconstruction}.
\newblock In \emph{Proceedings of the AAAI Conference on Artificial Intelligence}, volume~31.

\bibitem[{Van~der Maaten and Hinton(2008)}]{van2008visualizing}
Laurens Van~der Maaten and Geoffrey Hinton. 2008.
\newblock \href {https://www.jmlr.org/papers/volume9/vandermaaten08a/vandermaaten08a.pdf?fbcl} {Visualizing data using t-sne.}
\newblock \emph{Journal of machine learning research}, 9(11).

\bibitem[{Vilar et~al.(2023)Vilar, Freitag, Cherry, Luo, Ratnakar, and Foster}]{vilar2023prompting}
David Vilar, Markus Freitag, Colin Cherry, Jiaming Luo, Viresh Ratnakar, and George Foster. 2023.
\newblock \href {https://aclanthology.org/2023.acl-long.859/} {Prompting palm for translation: Assessing strategies and performance}.
\newblock In \emph{Proceedings of the 61st Annual Meeting of the Association for Computational Linguistics (Volume 1: Long Papers)}, pages 15406--15427.

\bibitem[{Wu et~al.(2022)Wu, Miwa, and Zhang}]{wu2022cognitive}
Yan~Jing Wu, Koji Miwa, and Haoyun Zhang. 2022.
\newblock \href {https://www.frontiersin.org/journals/psychology/articles/10.3389/fpsyg.2022.1055759/full} {Cognitive factors in bilingual language processing}.
\newblock \emph{Frontiers in Psychology}, 13.

\bibitem[{Zeng et~al.(2022)Zeng, Chen, and Guo}]{zeng2022first}
Tao Zeng, Chen Chen, and Jiashu Guo. 2022.
\newblock \href {https://www.frontiersin.org/journals/psychology/articles/10.3389/fpsyg.2022.986450/full} {First language translation involvement in second language word processing}.
\newblock \emph{Frontiers in Psychology}, 13:986450.

\bibitem[{Zhang et~al.(2023)Zhang, Li, Hauer, Shi, and Kondrak}]{zhang2023don}
Xiang Zhang, Senyu Li, Bradley Hauer, Ning Shi, and Grzegorz Kondrak. 2023.
\newblock \href {https://aclanthology.org/2023.emnlp-main.491/} {Don’t trust chatgpt when your question is not in english: A study of multilingual abilities and types of llms}.
\newblock In \emph{Proceedings of the 2023 Conference on Empirical Methods in Natural Language Processing}, pages 7915--7927.

\bibitem[{Zhu et~al.(2023)Zhu, Zimina, B{\'e}nard, Namdar, Ballier, Wisniewski, and Yun{\`e}s}]{zhu2023investigating}
Lichao Zhu, Maria Zimina, Maud B{\'e}nard, Behnoosh Namdar, Nicolas Ballier, Guillaume Wisniewski, and Jean-Baptiste Yun{\`e}s. 2023.
\newblock \href {https://aclanthology.org/2023.wmt-1.28/} {Investigating techniques for a deeper understanding of neural machine translation (nmt) systems through data filtering and fine-tuning strategies}.
\newblock In \emph{Proceedings of the Eighth Conference on Machine Translation}, pages 275--281.

\end{thebibliography}

\clearpage

\appendix

\section{Appendix}

\subsection{Implementation Details}
\label{appendix:details}

In this study, to minimize randomness introduced during the sampling process, we standardized the decoding method across all MLLMs to greedy decoding, which includes both Translator and Speaker LLMs. Furthermore, all MLLMs involved in the experiments are open-source models of the Instruct/Chat version: Phi-3-mini (3.8B) \footnote{\url{https://huggingface.co/microsoft/Phi-3-mini-128k-instruct}}, Phi-3-small (7B) \footnote{\url{https://huggingface.co/microsoft/Phi-3-small-128k-instruct}}, Gemma-1.1 (7B) \footnote{\url{https://huggingface.co/google/gemma-1.1-7b-it}}, Mistral-0.3 (7B) \footnote{\url{https://huggingface.co/mistralai/Mistral-7B-Instruct-v0.3}}, Llama-2 (7B) \footnote{\url{https://huggingface.co/meta-llama/Llama-2-7b-chat-hf}}, Qwen-1.5 (4B) \footnote{\url{https://huggingface.co/Qwen/Qwen1.5-4B-Chat}}, Qwen-2 (7B) \footnote{\url{https://huggingface.co/Qwen/Qwen2-7B-Instruct}}, and Qwen-1.5 (14B) \footnote{\url{https://huggingface.co/Qwen/Qwen1.5-14B-Chat}}.

At the same time, as we deployed Translator LLMs within NatLan that required designing translation prompts, we used GPT-4o \footnote{API version: \texttt{gpt-4o-2024-05-13}} to translate the dev sets of various disciplines in the C-Eval benchmark from Chinese to English. This ensures the quality of the translations in the prompts, with each discipline's dev set containing five examples, allowing us to construct five-shot translation prompts for each discipline. We also created five-shot Q\&A prompts using the C-Eval dev sets. In practical applications, we provide the MLLMs with prompts corresponding to the discipline currently being tested, thus maximizing the elicitation of their domain-specific knowledge. As for the MMMLU benchmark, we directly used official samples translated by human experts as examples in our prompts and employed GPT-4o-mini\footnote{API version: \texttt{gpt-4o-mini-2024-07-18}} as a cost-effective and powerful universal translator.

Since the Translator LLMs and Speaker LLMs used in the proposed NatLan method are required to undertake distinct processes, the former are required to translate questions from the target language to the native language, while the latter are required to provide answers based on the translated questions in the native language. Therefore, they use different sets of prompts. First, we report the details of the translation prompts used in our experiments as follows:\\
\noindent\rule [0.25\baselineskip] {\columnwidth} {1pt}
\texttt{<System Prompts>}\\
\texttt{You are a professional \{non-native language name\}-English translator. Translation rules: Proper nouns in English or \{non-native language name\} need to be translated according to the \{discipline name\} domain-specific terms, retain the original meaning to the greatest extent, and follow the original format in the translation process.}\\\\
\texttt{<Original Question Prompts>}\\
\texttt{Now help me translate the following sentence into English, only return the translated sentence, the original sentence is:\\ Question:\\\{original example[`question']\}\\Choices:\\A. \{original example[`choice A']\}\\B. \{original example[`choice B']\}\\C. \{original example[`choice C']\}\\D. \{original example[`choice D']\}\\Answer:}\\\\
\texttt{<Translated Question Prompts>}\\
\texttt{Question:\\\{translated example[`question']\}\\Choices:\\A. \{translated example[`choice A']\}\\B. \{translated example[`choice B']\}\\C. \{translated example[`choice C']\}\\D. \{translated example[`choice D']\}\\Answer:}\\
\noindent\rule [0.25\baselineskip] {\columnwidth} {1pt}

Furthermore, we report the details of the Q\&A prompts used in our experiments as follows:\\
\noindent\rule [0.25\baselineskip] {\columnwidth} {1pt}
\texttt{<System Prompts>}\\
\texttt{You are a professional \{discipline name\} expert, and you are currently answering a multiple-choice question about \{discipline name\}, you need to provide only one option as the answer based on the question, and you only need to return one single capital character as the answer.}\\\\
\texttt{<Question Prompts>}\\
\texttt{Question:\\\{translated example[`question']\}\\Choices:\\A. \{translated example[`choice A']\}\\B. \{translated example[`choice B']\}\\C. \{translated example[`choice C']\}\\D. \{translated example[`choice D']\}\\Answer:}\\\\
\texttt{<Answer Prompts>}\\
\texttt{\{example[`answer']\}}\\
\noindent\rule [0.25\baselineskip] {\columnwidth} {1pt}

Additionally, the disciplines used for evaluating the advantages of DT retention, discussed in \S \ref{sec:DT_comp}, encompass ten domains rich in domain-specific terms, including Anatomy, Astronomy, College Biology, College Chemistry, College Computer Science, College Medicine, High School Biology, High School Chemistry, High School Computer Science, and Virology.

\subsection{Comparative Analysis of Chinese-to-English Translation Cases}
\label{appendix:cases}

\begin{table*}[ht]
\centering
\resizebox{2\columnwidth}{!}{
\begin{tabular}{m{0.24\textwidth}m{0.3\textwidth}m{0.3\textwidth}m{0.16\textwidth}}
\hline
\textbf{Original Question} & \textbf{Google-MT Trans. Question} & \textbf{NatLan Trans. Question} & \textbf{Answers} \\ 
\hline
\begin{tabular}{@{}p{0.24\textwidth}@{}}
\begin{CJK*}{UTF8}{gbsn}
某计算机的指令系统中共有101条不同的指令，采用微程序控制方式时，控制存储器中具有的微程序数目至少是$\_\_\_\_$。
\end{CJK*}\\
\begin{CJK*}{UTF8}{gbsn}
A. 100
\end{CJK*}\\
\begin{CJK*}{UTF8}{gbsn}
B. 102
\end{CJK*}\\
\begin{CJK*}{UTF8}{gbsn}
C. 103
\end{CJK*}\\
\begin{CJK*}{UTF8}{gbsn}
D. 104
\end{CJK*}
\end{tabular}
& \begin{tabular}{@{}p{0.3\textwidth}@{}}
There are 101 different instructions in the instruction system of a certain computer. When using microprogram control, the number of microprograms in the control memory is at least $\_\_\_\_$.\\
A. 100\\
B. 102\\
C. 103\\
D. 104\\
\end{tabular}
& \begin{tabular}{@{}p{0.3\textwidth}@{}}
In a computer's instruction set with a total of 101 different instructions, the minimum number of microprograms \textcolor{darkgreen}{required} in the control memory when using microprogram control is $\_\_\_\_$.\\
A. 100\\
B. 102\\
C. 103\\
D. 104\\
\end{tabular}
& \begin{tabular}{@{}p{0.16\textwidth}@{}}
\\
Original: \textcolor{darkgreen}{B}\\\\
+Google-MT: \textcolor{darkred}{C}\\\\
+NatLan : \textcolor{darkgreen}{B}\\\\
True Label : \textcolor{darkgreen}{B}
\end{tabular} \\ 
\hline
\begin{tabular}{@{}p{0.24\textwidth}@{}}
\begin{CJK*}{UTF8}{gbsn}
迁都后对帕朗卡拉亚的影响有$\_\_\_\_$。
\end{CJK*}\\
\begin{CJK*}{UTF8}{gbsn}
A. 有利于缓解住房紧张问题
\end{CJK*}\\
\begin{CJK*}{UTF8}{gbsn}
B. 有利于缓解交通拥堵状况
\end{CJK*}\\
\begin{CJK*}{UTF8}{gbsn}
C. 有利于环境污染的治理
\end{CJK*}\\
\begin{CJK*}{UTF8}{gbsn}
D. 基础设施的完善
\end{CJK*}
\end{tabular}
& \begin{tabular}{@{}p{0.3\textwidth}@{}}
The impact of the capital relocation on Palangkaraya is$\_\_\_\_$.\\
A. It is conducive to alleviating housing shortages\\
B. It is conducive to alleviating traffic congestion\\
C. It is conducive to the control of environmental pollution\\
D. The improvement of infrastructure
\end{tabular}
& \begin{tabular}{@{}p{0.3\textwidth}@{}}
The impact of the capital relocation on Palangkaraya would \textcolor{darkgreen}{include}$\_\_\_\_$.\\
A. Alleviating housing shortages\\
B. Alleviating traffic congestion\\
C. Facilitating environmental pollution control\\
D. Improvement of infrastructure
\end{tabular}
& \begin{tabular}{@{}p{0.16\textwidth}@{}}
\\
Original: \textcolor{darkred}{A}\\\\
+Google-MT: \textcolor{darkred}{C}\\\\
+NatLan : \textcolor{darkgreen}{D}\\\\
True Label: \textcolor{darkgreen}{D}
\end{tabular} \\ 
\hline
\begin{tabular}{@{}p{0.24\textwidth}@{}}
\begin{CJK*}{UTF8}{gbsn}
下列各物质完全燃烧，产物除二氧化碳和水外，还有其他物质的是$\_\_\_\_$。
\end{CJK*}\\
\begin{CJK*}{UTF8}{gbsn}
A. 甲烷
\end{CJK*}\\
\begin{CJK*}{UTF8}{gbsn}
B. 乙烯
\end{CJK*}\\
\begin{CJK*}{UTF8}{gbsn}
C. 氯乙烯
\end{CJK*}\\
\begin{CJK*}{UTF8}{gbsn}
D. 乙醇
\end{CJK*}
\end{tabular}
& \begin{tabular}{@{}p{0.3\textwidth}@{}}
When the following substances are completely \textcolor{darkred}{burned}, the products include carbon dioxide and water, and other substances$\_\_\_\_$.\\
A. Methane\\
B. Ethylene\\
C. Vinyl chloride\\
D. Ethanol
\end{tabular}
& \begin{tabular}{@{}p{0.3\textwidth}@{}}
Among the following substances, which one, when completely \textcolor{darkgreen}{combusted}, produces products other than carbon dioxide and water?$\_\_\_\_$.\\
A. Methane\\
B. Ethylene\\
C. Vinyl chloride\\
D. Ethanol
\end{tabular}
& \begin{tabular}{@{}p{0.16\textwidth}@{}}
\\
Original: \textcolor{darkgreen}{C}\\\\
+Google-MT: \textcolor{darkred}{D}\\\\
+NatLan : \textcolor{darkgreen}{C}\\\\
True Label : \textcolor{darkgreen}{C}
\end{tabular} \\ 
\hline
\begin{tabular}{@{}p{0.24\textwidth}@{}}
\begin{CJK*}{UTF8}{gbsn}
下列有关$NaHCO_3$与\end{CJK*}\\
\begin{CJK*}{UTF8}{gbsn}$Na_2CO_3$的说法中不正确的是$\_\_\_\_$。
\end{CJK*}\\
\begin{CJK*}{UTF8}{gbsn}
A. 在水中溶解性：
\end{CJK*}
$Na_2CO_3 < NaHCO_3$\\
\begin{CJK*}{UTF8}{gbsn}
B. 与相同浓度酸反应的剧烈程度：
\end{CJK*}
$Na_2CO_3 < NaHCO_3$\\
\begin{CJK*}{UTF8}{gbsn}
C. 热稳定性：
\end{CJK*}
$Na_2CO_3 < NaHCO_3$\\
\begin{CJK*}{UTF8}{gbsn}
D. 二者间在一定条件下可相互转化
\end{CJK*}
\end{tabular}
& \begin{tabular}{@{}p{0.3\textwidth}@{}}
Which of the following statements about $NaHCO_3$ and $Na_2CO_3$ is incorrect$\_\_\_\_$.\\
A. Solubility in water: \\$Na_2CO_3 < NaHCO_3$\\
B. \textcolor{darkred}{The intensity of the reaction} with the same concentration of acid: $Na_2CO_3 < NaHCO_3$\\
C. Thermal stability: \\$Na_2CO_3 < NaHCO_3$\\
D. The two can be converted into each other under certain conditions
\end{tabular}
& \begin{tabular}{@{}p{0.3\textwidth}@{}}
Which of the following statements about $NaHCO_3$ and $Na_2CO_3$ is incorrect?$\_\_\_\_$.\\
A. Solubility in water: \\$Na_2CO_3 < NaHCO_3$\\
B. \textcolor{darkgreen}{Reactivity} with equal concentration acids: $Na_2CO_3 < NaHCO_3$\\
C. Thermal stability: \\$Na_2CO_3 < NaHCO_3$\\
D. They can transform into each other under certain conditions
\end{tabular}
& \begin{tabular}{@{}p{0.16\textwidth}@{}}
\\\\
Original: \textcolor{darkred}{B}\\\\
+Google-MT: \textcolor{darkred}{B}\\\\
+NatLan : \textcolor{darkgreen}{C}\\\\
True Label : \textcolor{darkgreen}{C}
\end{tabular} \\ 
\hline
\end{tabular}
}
\caption{Supplementary Chinese-to-English translation cases, with cases sampled from the C-Eval test sets. The Speaker LLM is Phi-3-mini (3.8B) and the Translator LLM is Qwen-1.5 (14B) for this case study.}
\label{tab:appendix-case}
\end{table*}

As a supplement to Table \ref{tab:case-study}, we report a more detailed comparative analysis of Chinese-to-English translation cases between Google-MT and the proposed NatLan in Table \ref{tab:appendix-case}.

As shown in Table \ref{tab:appendix-case}, in the examples from the first two rows, NatLan provides more explicit operation descriptions in translations, enabling Speaker LLMs to more easily understand the relationship between the domain-specific entities in the questions and answer candidates. In the cases presented in the latter two rows, NatLan delivers translations with more accurate domain-specific entity descriptions. For these two questions pertaining to the High School Chemistry discipline, the enriched semantic comprehension of the Translator LLMs enables NatLan to generate terms that aligns more closely with domain-specific usage. For instance, it translates to \underline{``combusted''}, which is preferred in chemical contexts, rather than the general term \underline{``burned''}, and \underline{``Reactivity''} instead of \underline{``The intensity of reaction''}.

\subsection{More Cases in Knowledge Activation}
\label{appendix:act}

\begin{figure*}[ht] 
\centering
\includegraphics[width=1\linewidth]{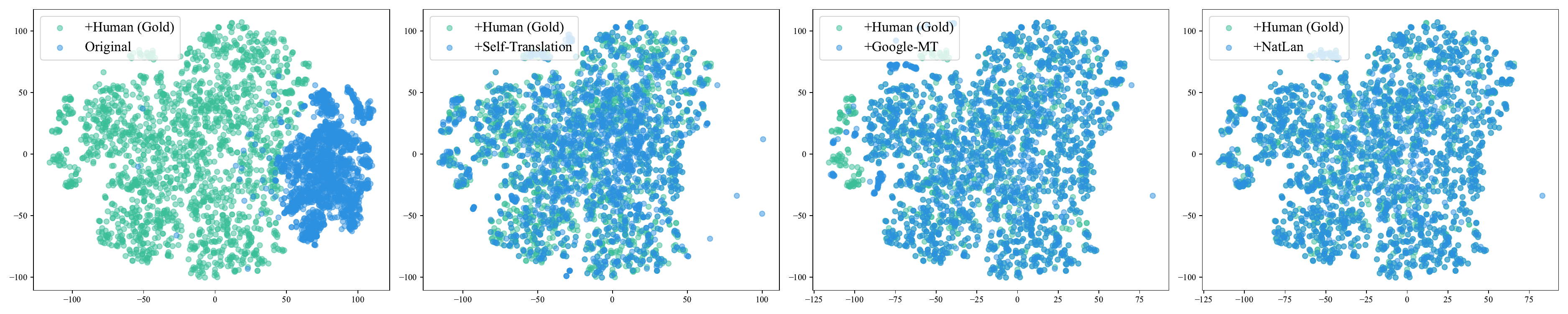}
\caption{The visualized knowledge activation distributions on the Arabic version of the MMMLU benchmark, with the Speaker LLM: Phi-3-small (7B). The greater the overlap with green (human gold standard), the more accurate the knowledge activation is considered.}
\label{act_analysis_supp-1}
\end{figure*}

\begin{figure*}[ht] 
\centering
\includegraphics[width=1\linewidth]{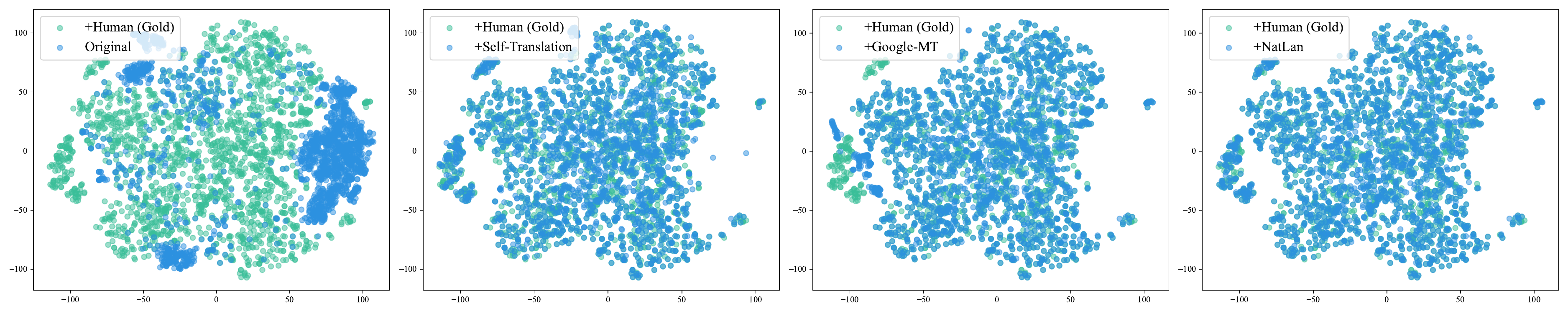}
\caption{The visualized knowledge activation distributions on the Chinese version of the MMMLU benchmark, with the Speaker LLM: Phi-3-small (7B). The greater the overlap with green (human gold standard), the more accurate the knowledge activation is considered.}
\label{act_analysis_supp-2}
\end{figure*}

\begin{figure*}[ht] 
\centering
\includegraphics[width=1\linewidth]{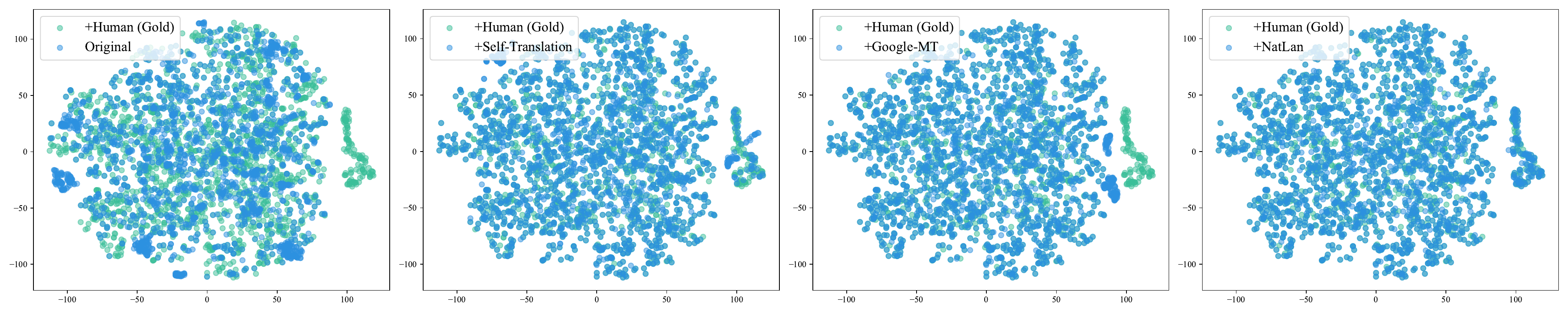}
\caption{The visualized knowledge activation distributions on the German version of the MMMLU benchmark, with the Speaker LLM: Phi-3-small (7B). The greater the overlap with green (human gold standard), the more accurate the knowledge activation is considered.}
\label{act_analysis_supp-3}
\end{figure*}

\begin{figure*}[ht] 
\centering
\includegraphics[width=1\linewidth]{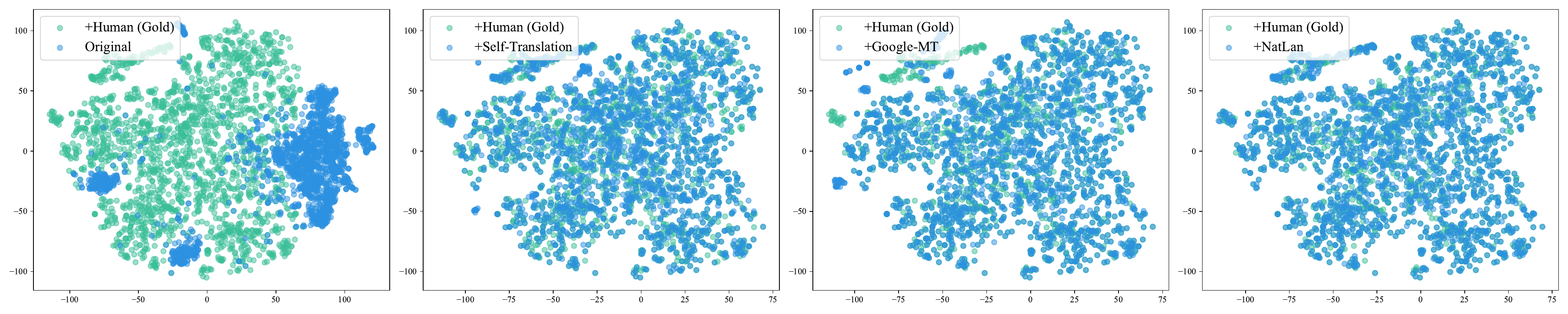}
\caption{The visualized knowledge activation distributions on the Japanese version of the MMMLU benchmark, with the Speaker LLM: Phi-3-small (7B). The greater the overlap with green (human gold standard), the more accurate the knowledge activation is considered.}
\label{act_analysis_supp-4}
\end{figure*}

As a supplement to \S \ref{sec:act_analysis}, we report the visualized knowledge activation distributions for the Arabic, Chinese, German, Japanese version of the MMMLU benchmark, with the Speaker LLM: Phi-3-small (7B), in Figure \ref{act_analysis_supp-1}, Figure \ref{act_analysis_supp-2}, Figure \ref{act_analysis_supp-3} and Figure \ref{act_analysis_supp-4}, respectively. Moreover, we report the case used to measure differences in knowledge activation in this experiment, which were sampled from the C-Eval val/test sets. Detailed content is shown in Table \ref{tab:appendix-act-case}.

\begin{table*}[ht]
\centering
\resizebox{2\columnwidth}{!}{
\begin{tabular}{m{0.24\textwidth}m{0.3\textwidth}m{0.3\textwidth}m{0.16\textwidth}}
\hline
\textbf{Original Question} & \textbf{Google-MT Trans. Question} & \textbf{NatLan Trans. Question} & \textbf{Answers} \\ 
\hline
\begin{tabular}{@{}p{0.24\textwidth}@{}}
\begin{CJK*}{UTF8}{gbsn}
间址寻址第一次访问内存所得到的信息经$\_\_\_\_$传送到MDR。
\end{CJK*}\\
\begin{CJK*}{UTF8}{gbsn}
A. 数据总线
\end{CJK*}\\
\begin{CJK*}{UTF8}{gbsn}
B. 地址总线
\end{CJK*}\\
\begin{CJK*}{UTF8}{gbsn}
C. 控制总线
\end{CJK*}\\
\begin{CJK*}{UTF8}{gbsn}
D. 总线控制器
\end{CJK*}
\end{tabular}
& \begin{tabular}{@{}p{0.3\textwidth}@{}}
The information obtained by indirect addressing when accessing the memory for the first time is transmitted to MDR via$\_\_\_\_$.\\
A. Data bus\\
B. Address bus\\
C. Control bus\\
D. Bus controller
\end{tabular}
& \begin{tabular}{@{}p{0.3\textwidth}@{}}
The information obtained from the first memory access using indirect addressing is transmitted to the MDR via$\_\_\_\_$.\\
A. data bus\\
B. address bus\\
C. control bus\\
D. bus controller
\end{tabular}
& \begin{tabular}{@{}p{0.16\textwidth}@{}}
\\
Original: \textcolor{darkred}{B}\\\\
+Google-MT: \textcolor{darkred}{B}\\\\
+NatLan : \textcolor{darkgreen}{A}\\\\
True Label : \textcolor{darkgreen}{A}\\\\
\end{tabular} \\ 
\hline
\end{tabular}
}
\caption{Cases sampled from the C-Eval val/test sets for knowledge activation analysis in \S \ref{sec:act_analysis}.}
\label{tab:appendix-act-case}
\end{table*}

\subsection{Overall Performance Results on the C-Eval benchmark}
\label{appendix:c-eval}
As shown in Table \ref{tab: comparison}, our proposed NatLan also achieved optimal performance on the C-Eval benchmark. However, unlike MMMLU, the C-Eval benchmark does not have a human expert-constructed gold standard English version and only includes Chinese. Therefore, we primarily used it for ablation studies and case studies.

\begin{table}[ht]
\centering
\resizebox{1\columnwidth}{!}{
\begin{tabular}{lccc}
\hline
\textbf{Model} & \textbf{Lang.} & \textbf{Avg.} & \textbf{Avg. (Hard)} \\ 
\hline
Phi-3-mini (3.8B) & zh & 41.2 & 36.3\\
\rowcolor{yellow!10}
+Self-Translation & en & 43.8 & 37.7\\
\rowcolor{yellow!10}
+Google-MT & en & 50.9 & 40.4\\
\rowcolor{green!10}
+NatLan & en & 51.3 & 41.3\\
\hline
Phi-3-small (7B) & zh & 49.0 & 41.6\\
\rowcolor{yellow!10}
+Self-Translation & en & 52.0 & 42.1\\
\rowcolor{yellow!10}
+Google-MT & en & 55.7 & 42.7\\
\rowcolor{green!10}
+NatLan & en & 55.9 & 44.7\\
\hline
Gemma-1.1 (7B) & zh & 44.4 & 36.3\\
\rowcolor{red!10}
+Self-Translation & en & 41.9 & 33.9\\
\rowcolor{yellow!10}
+Google-MT & en & 46.7 & 38.2\\
\rowcolor{green!10}
+NatLan & en & 47.7 & 38.6\\
\hline
Mistral-0.3 (7B) & zh & 42.8 & 32.6\\
\rowcolor{red!10}
+Self-Translation & en & 34.8 & 30.9\\
\rowcolor{yellow!10}
+Google-MT & en & 48.0 & 33.3\\
\rowcolor{green!10}
+NatLan & en & 48.4 & 35.3\\
\hline
Llama-2 (7B) & zh & 21.3 & 14.7\\
\rowcolor{red!10}
+Self-Translation & en & 9.6 & 10.3\\
\rowcolor{yellow!10}
+Google-MT & en & 25.4 & 15.1\\
\rowcolor{green!10}
+NatLan & en & 27.6 & 18.6\\
\hline
\end{tabular}
}
\caption{Performance on the C-Eval Chinese benchmark, measured by accuracy. \textit{Lang.} indicates the language of the questions. The meanings of these colors are the same as in Table \ref{tab: mmmlu-all}. NatLan employs Qwen MLLMs as the Chinese-to-English translators, selecting the Qwen configurations reported as optimal in \S \ref{sec:ablation_1}.}
\label{tab: comparison}
\end{table}

\subsection{Analysis of NatLan with Different Translators in Various Domains}
\label{appendix:ablation}

\begin{figure*}[ht] 
\centering
\includegraphics[width=0.9\linewidth]{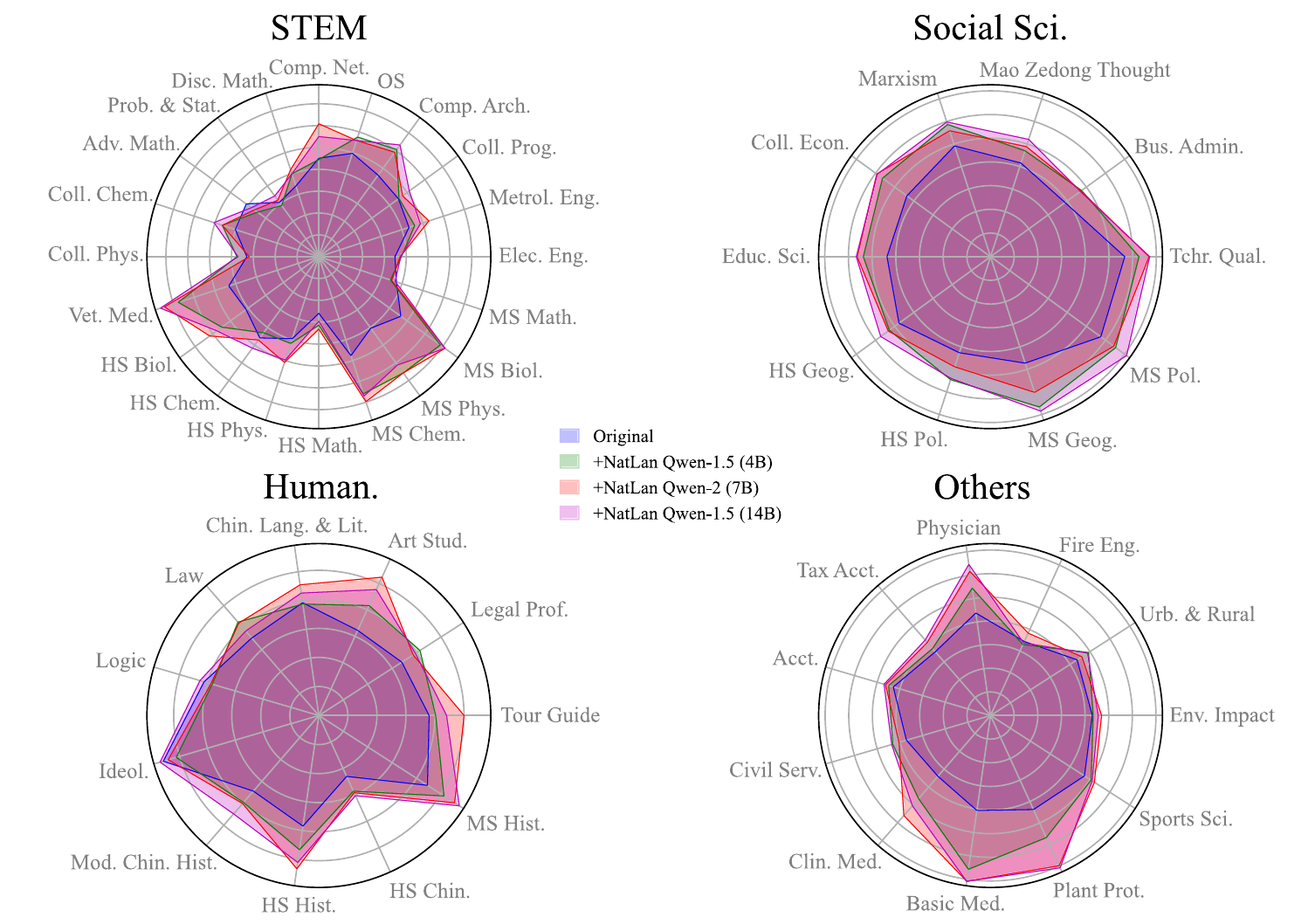}
\caption{Performance comparison of NatLan using different Translator LLMs in the C-Eval test sets, divided into 52 distinct disciplines, with Phi-3-mini (3.8B) as the Speaker LLMs.}
\label{phi3_mini_radar}
\end{figure*}

\begin{figure*}[ht] 
\centering
\includegraphics[width=0.9\linewidth]{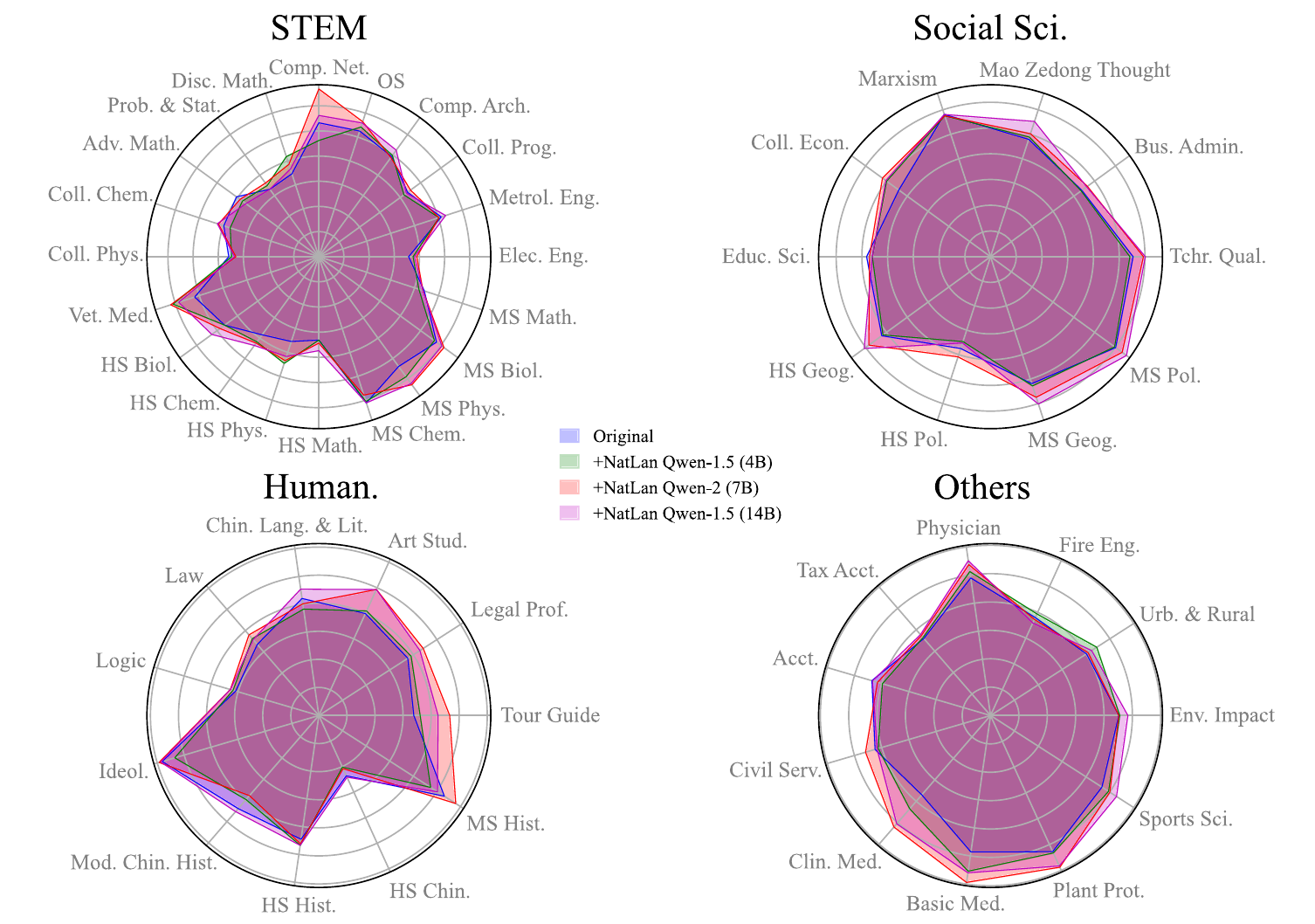}
\caption{Performance comparison of NatLan using different Translator LLMs in the C-Eval test sets, divided into 52 distinct disciplines, with Gemma-1.1 (7B) as the Speaker LLMs.}
\label{gemma_radar}
\end{figure*}

\begin{figure*}[ht] 
\centering
\includegraphics[width=0.9\linewidth]{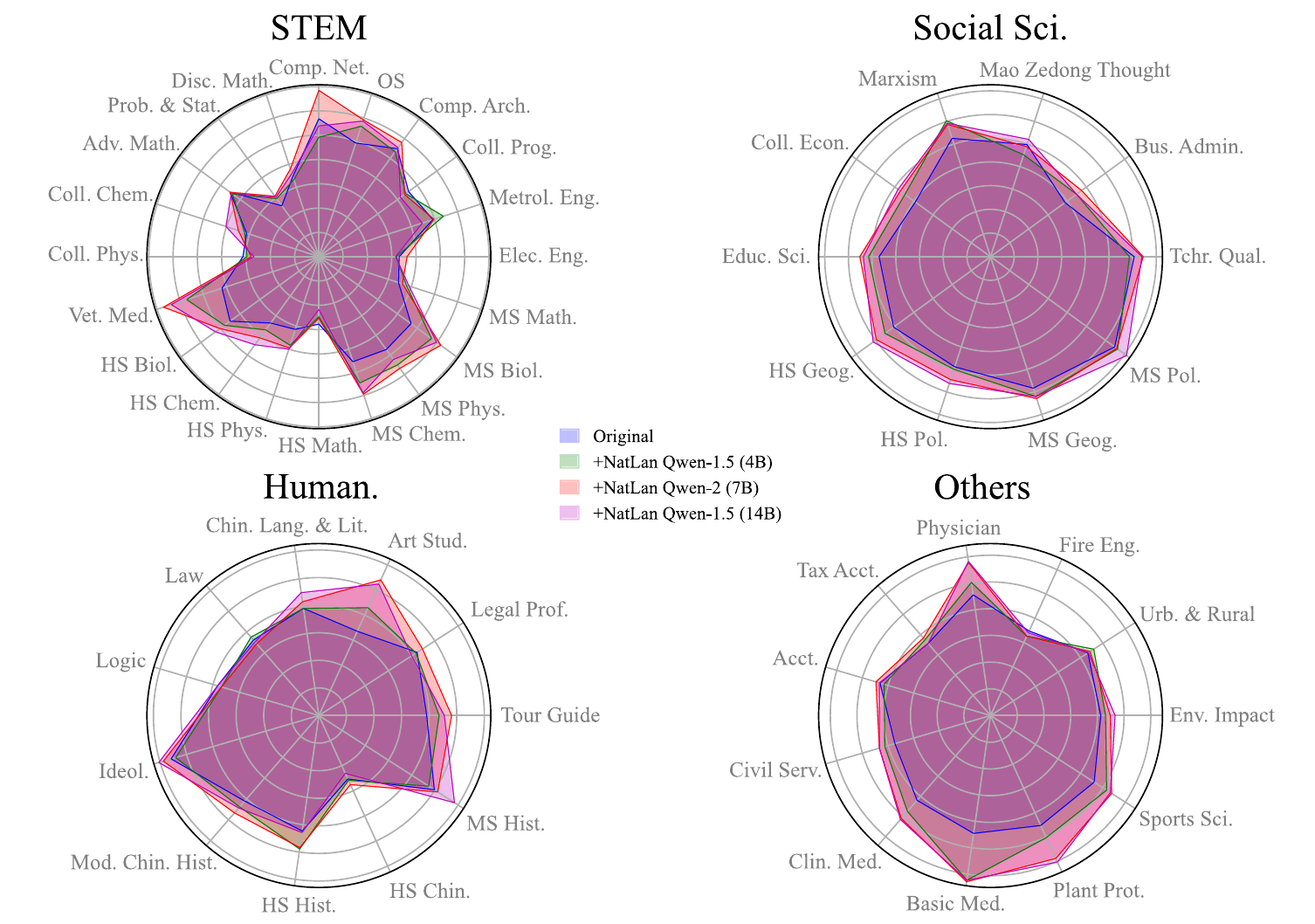}
\caption{Performance comparison of NatLan using different Translator LLMs in the C-Eval test sets, divided into 52 distinct disciplines, with Mistral-0.3 (7B) as the Speaker LLMs.}
\label{mistral_radar}
\end{figure*}

\begin{figure*}[ht] 
\centering
\includegraphics[width=0.9\linewidth]{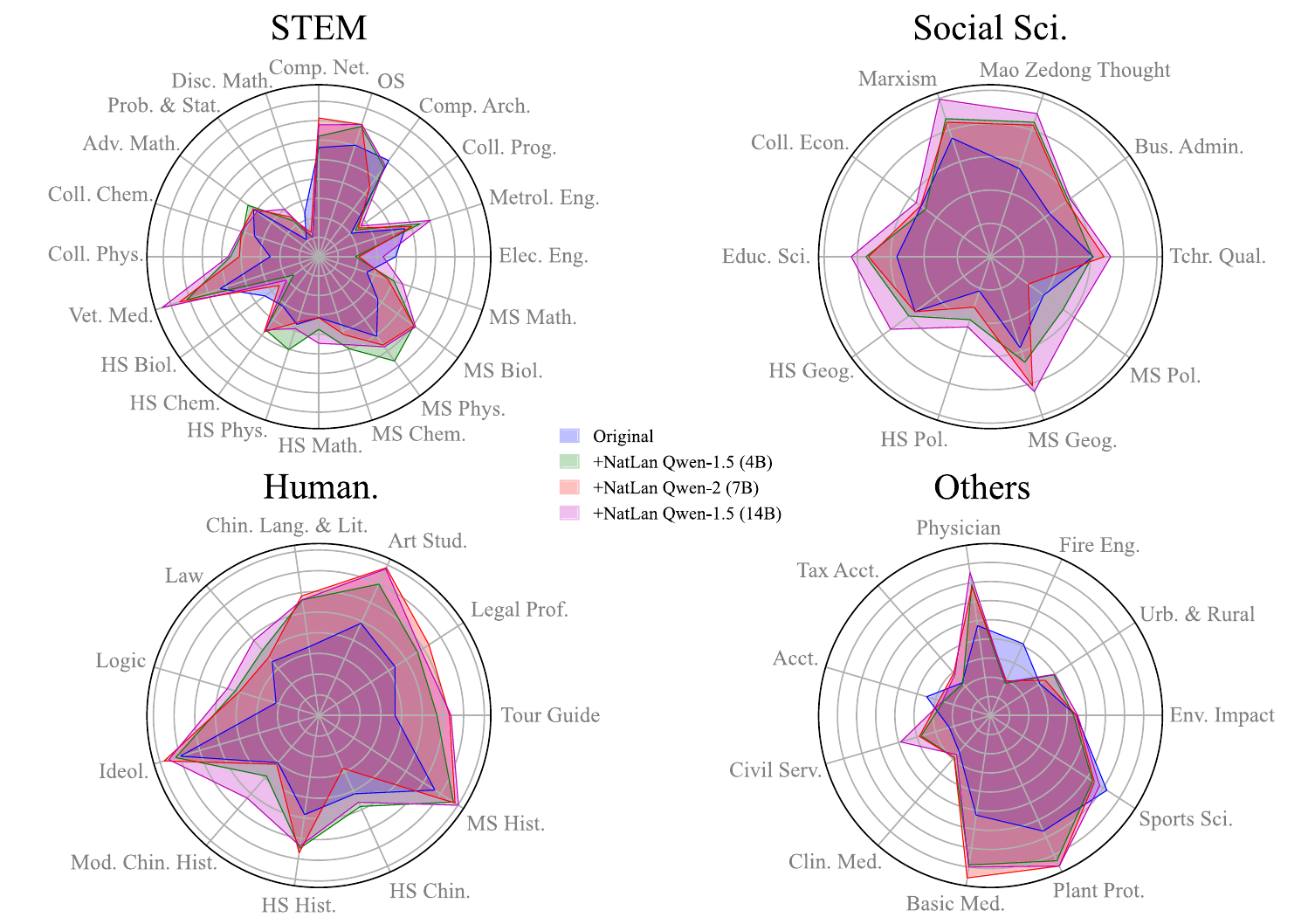}
\caption{Performance comparison of NatLan using different Translator LLMs in the C-Eval test sets, divided into 52 distinct disciplines, with Llama-2 (7B) as the Speaker LLMs.}
\label{llama2_radar}
\end{figure*}

As a supplement to Figure \ref{phi3_radar}, we present a detailed performance analysis of NatLan, employing three different Translator LLMs applied to various Speaker LLMs, across specific disciplines. These include Phi-3-mini (3.8B) in Figure \ref{phi3_mini_radar}, Gemma-1.1 (7B) in Figure \ref{gemma_radar}, Mistral-0.3 (7B) in Figure \ref{mistral_radar}, and Llama-2 (7B) in Figure \ref{llama2_radar}.

As shown in these figures, NatLan has provided widespread and consistent performance improvements across all Speaker LLMs, with only minor performance declines in a very few disciplines. Furthermore, across each Speaker LLM, performance improvements and the disciplines where declines occur vary due to differences in performance preferences, the proportion of different language data in the training corpora, and variations in data sources and quality. This variation highlights that the knowledge elicitation facilitated by NatLan, aside from the influence of Translator LLMs, is primarily dependent on the capabilities of the Speaker LLMs in their native languages.

Additionally, it is important to note that since NatLan relies heavily on the collaboration of MLLMs, it also demands a high level of compliance with instructions from the MLLMs. As shown in Figure \ref{llama2_radar}, Llama-2 (7B), compared to other Speaker LLMs, has relatively weaker instruction-following capabilities. Consequently, it is more prone to producing answers that do not conform to the prescribed format during testing. We applied a strict evaluation criterion in these instances, considering any output that did not meet the established format as incorrect. Thus, the performance improvements brought about by NatLan using different Translator LLMs on Llama-2 (7B) show relatively greater variability. However, from a holistic perspective, disregarding the variations between different Translator LLMs, NatLan still manages to provide stable performance improvements for Llama-2 (7B). This further confirms the superiority of the proposed NatLan method.

Furthermore, we have reported the detailed performance evaluation scores of NatLan and top-notch related methods in Table \ref{tab: appendix-all} for all settings, as a supplement to Table \ref{tab: comparison} and Table \ref{tab: ablation_natlan}

\subsection{NatLan Produces More Relative Improvements} \label{appendix:relative}

To explore in more depth, we conducted a detailed performance analysis of Google-MT and our proposed NatLan method on the validation sets of specific disciplines within the C-Eval benchmark.

We define our analysis process as follows: Considering each discipline individually, we calculate the relative performance improvements brought by NatLan/Google-MT compared to having Speaker LLMs directly answer questions in Chinese (Original).   Specifically, this involves computing the relative increase in the number of correct answers provided by NatLan/Google-MT compared to the Original.   Subsequently, we apply Min-Max Normalization to the relative improvements achieved by NatLan/Google-MT across various disciplines, resulting in normalized relative improvements.

\begin{figure}[ht] 
\centering
\includegraphics[width=1\linewidth]{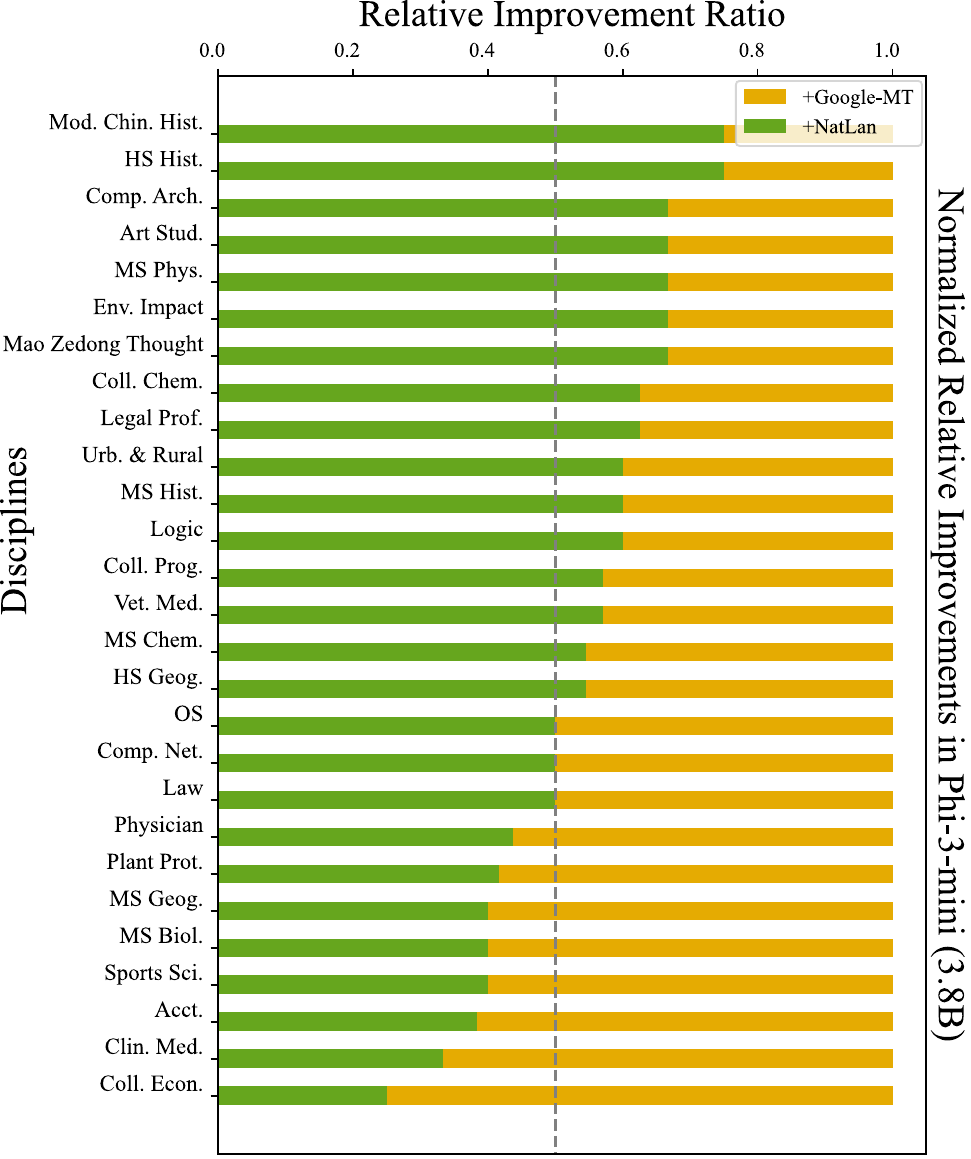}
\caption{Normalized relative improvements in specific disciplines, with the dashed grey line indicating where their respective relative improvements are equivalent.}
\label{phi3_stacked_bar}
\end{figure}

As shown in Figure \ref{phi3_stacked_bar}, NatLan provides more relative improvements than Google-MT in the majority of disciplines. It is important to note that we have excluded disciplines from this analysis where neither method provided more correct answers than the Original. Additionally, since the performance gains from Self-Translation are quite limited and often result in frequent performance declines, this method has not been included in the analysis.

\begin{table*}[ht]
\centering
\resizebox{2\columnwidth}{!}{
\begin{tabular}{lccccccc}
\hline
\textbf{Model} & \textbf{Lang.} & \textbf{STEM} & \textbf{Social Sci.} & \textbf{Human.} & \textbf{Others} & \textbf{Avg.} & \textbf{Avg. (Hard)} \\ 
\hline
\multicolumn{8}{c}{\textit{Translator LLMs}}\\
\hline
Qwen-1.5 (4B) & zh & 55.2 & 73.7 & 62.0 & 54.9 & 60.1 & 42.3\\
Qwen-2 (7B) & zh & 71.4 & 88.7 & 80.9 & 81.8 & 78.9 & 56.7\\
Qwen-1.5 (14B) & zh & 69.9 & 86.7 & 76.3 & 71.6 & 74.9 & 58.9\\
\hline
\multicolumn{8}{c}{\textit{Speaker LLMs}}\\
\hline
Phi-3-mini (3.8B) & zh & 40.5 & 46.9 & 37.8 & 40.5 & 41.2 & 36.3\\
\rowcolor{yellow!10}
+Self-Translation & en & 44.8 & 48.9 & 37.4 & 43.7 & 43.8 & 37.7\\
\rowcolor{yellow!10}
+Google-MT & en & 50.1 & 56.3 & 46.7 & 51.4 & 50.9 & 40.4\\
+NatLan Qwen-1.5 (4B) & en & 47.6 & 56.5 & 41.8 & 47.7 & 48.1 & 37.9\\
+NatLan Qwen-2 (7B) & en & 50.5 & 56.1 & 45.4 & 51.7 & 50.8 & 39.9\\
\rowcolor{green!10}
+NatLan Qwen-1.5 (14B) & en & 50.6 & 59.2 & 45.1 & 51.7 & 51.3 & 41.3\\
\hline
Phi-3-small (7B) & zh & 47.9 & 57.7 & 43.4 & 48.8 & 49.0 & 41.6\\
\rowcolor{yellow!10}
+Self-Translation & en & 51.4 & 59.6 & 46.4 & 51.8 & 52.0 & 42.1\\
\rowcolor{yellow!10}
+Google-MT & en & 54.0 & 63.5 & 51.0 & 56.5 & 55.7 & 42.7\\
+NatLan Qwen-1.5 (4B) & en & 51.8 & 60.5 & 47.8 & 52.1 & 52.7 & 41.9\\
+NatLan Qwen-2 (7B) & en & 54.1 & 64.6 & 50.5 & 57.1 & 56.0 & 43.5\\
\rowcolor{green!10}
+NatLan Qwen-1.5 (14B) & en & 54.3& 63.4 & 51.6 & 56.4 & 55.9 & 44.7\\
\hline
Gemma-1.1 (7B) & zh & 44.6 & 49.9 & 40.1 & 43.6 & 44.4 & 36.3\\
\rowcolor{red!10}
+Self-Translation & en & 42.3 & 44.9 & 38.2 & 42.3 & 41.9 & 33.9\\
\rowcolor{yellow!10}
+Google-MT & en & 47.5 & 50.4 & 41.9 & 46.5 & 46.7 & 38.2\\
+NatLan Qwen-1.5 (4B) & en & 45.5 & 49.9 & 39.1 & 45.4 & 45.0 & 38.2\\
\rowcolor{green!10}
+NatLan Qwen-2 (7B) & en & 47.5 & 53.3 & 43.0 & 47.5 & 47.7 & 38.6\\
+NatLan Qwen-1.5 (14B) & en & 47.1 & 53.7 & 43.1 & 47.5 & 47.6 & 38.0\\
\hline
Mistral-0.3 (7B) & zh & 40.5 & 51.1 & 40.3 & 41.7 & 42.8 & 32.6\\
\rowcolor{red!10}
+Self-Translation & en & 35.5 & 36.1 & 31.6 & 35.6 & 34.8 & 30.9\\
\rowcolor{yellow!10}
+Google-MT & en & 44.5 & 55.9 & 45.8 & 49.2 & 48.0 & 33.3\\
+NatLan Qwen-1.5 (4B) & en & 43.4 & 53.9 & 42.0 & 45.8 & 45.6 & 33.6\\
\rowcolor{green!10}
+NatLan Qwen-2 (7B) & en & 46.5 & 56.5 & 44.7 & 48.4 & 48.4 & 35.3\\
+NatLan Qwen-1.5 (14B) & en & 44.8 & 57.3 & 44.1 & 48.4 & 47.8 & 35.5\\
\hline
Llama-2 (7B) & zh & 18.9 & 25.9 & 21.6 & 20.9 & 21.3 & 14.7\\
\rowcolor{red!10}
+Self-Translation & en & 8.7 & 8.7 & 11.5 & 9.6 & 9.6 & 10.3\\
\rowcolor{yellow!10}
+Google-MT & en & 19.9 & 31.9 & 29.9 & 24.9 & 25.4 & 15.1\\
+NatLan Qwen-1.5 (4B) & en & 22.3 & 31.8 & 28.4 & 23.2 & 25.6 & 18.7\\
+NatLan Qwen-2 (7B) & en & 21.4 & 30.8 & 28.3 & 24.0 & 25.2 & 17.3\\
\rowcolor{green!10}
+NatLan Qwen-1.5 (14B) & en & 23.3 & 36.3 & 30.4 & 24.8 & 27.6 & 18.6\\
\hline
\end{tabular}
}
\caption{Detailed performance scores (accuracy) of NatLan and top-notch related methods under different configurations on the C-Eval test sets. The meanings assigned to the different colors correspond to those in Table \ref{tab: comparison}.}
\label{tab: appendix-all}
\end{table*}

\end{document}